\documentclass[]{article}

\usepackage{lineno,hyperref}
\usepackage[T1]{fontenc}
\usepackage[utf8]{inputenc}
\hypersetup{colorlinks = false,linkcolor = black,anchorcolor =red,citecolor = black,filecolor =red,urlcolor = red, pdfauthor=author}
\usepackage{etoolbox}
\usepackage{amsmath}
\usepackage{amssymb}
\usepackage{subcaption}
\usepackage{multirow}
\usepackage{booktabs}
\usepackage{float}
\usepackage{siunitx}
\usepackage{algorithm}
\usepackage[noend]{algpseudocode}
\usepackage{soul}
\usepackage{adjustbox}
\usepackage[affil-it]{authblk}

\makeatletter
\def\ps@pprintTitle{%
 \let\@oddhead\@empty
 \let\@evenhead\@empty
 \def\@oddfoot{}%
 \let\@evenfoot\@oddfoot}
\makeatother

\modulolinenumbers[5]
\DeclareUnicodeCharacter{0096}{}

\newcommand{\vek}[1]{\mathchoice{\displaystyle\boldsymbol#1}
{\textstyle\boldsymbol#1}{\scriptstyle\boldsymbol#1}
{\scriptscriptstyle\boldsymbol#1}}

\usepackage[nopostdot,nonumberlist,acronym,toc,section]{glossaries}
\glsdisablehyper

\newcommand{\spatialdomain}{\mathcal{G}}
\newcommand{\samplespace}{\varOmega}
	
\newcommand{\reftab}[1]{Table~\ref{#1}}
\newcommand{\reffig}[1]{Fig.~\ref{#1}}

\newcommand{\refeq}[1]{Eq.~(\ref{#1})}

\title{Long short-term relevance learning}

\author[1,2,*]{\small B.P. van de Weg}
\author[1]{\small L. Greve}
\author[2]{\small B. Rosic}

\affil[1]{\small Volkswagen AG, Group Innovation, D-38436 Wolfsburg, Germany}
\affil[2]{Applied Mechanics and Data Analysis, Faculty of Engineering Technology, University of Twente, P.O. Box 217, 7500 AE Enschede, The Netherlands}
\affil[*]{Corresponding author: B.P. van de Weg, bram.van.de.weg@volkswagen.de}
\usepackage{a4wide}
\begin{document}
\date{}
\maketitle

\renewcommand{\thefootnote}{\arabic{footnote}}
\begin{abstract}
To incorporate prior knowledge as well as measurement uncertainties in the traditional long short term memory (LSTM) neural networks, an efficient sparse Bayesian training algorithm is introduced to the network architecture. The proposed scheme automatically determines relevant neural connections and adapts accordingly, in contrast to the classical LSTM solution.  Due to its flexibility, the new LSTM scheme is less prone to overfitting, and hence can approximate time dependent solutions by use of a smaller data set. On a structural nonlinear finite element application we show that the self-regulating framework does not require prior knowledge of a suitable network architecture and size, while ensuring satisfying accuracy at reasonable computational cost.
\end{abstract}

\noindent \textit{keywords: }LSTM, Neural network, Automatic relevance determination, Bayesian, Sparsity, Finite element model

\section{Introduction}
Recent trends in automotive industry have steered the rapid development of CAE as an important tool in reducing the overall production cost and lead time. Although tremendously useful, CAE predictive capabilities are negatively affected by its extensive computational cost. High-fidelity simulations like FEM analyses have failed to address quick-pace engineering decisions. The usage of computationally more efficient surrogate models help solving this dilemma.

Much uncertainty, however, still exists about the choice and design of an optimal proxy model, especially when considered in a highly nonlinear setting, e.g. in nonlinear mechanical applications. Up to date many approaches to quantify uncertainty have been investigated, examples of which are autoregressive–moving-average (ARMA) \cite{P.Whittle.1951}, support vector machine (SVM) \cite{Boser.1992}, Gaussian process (GP) \cite{Rasmussen.2005}, radial basis function (RBF) \cite{Broomhead.1988} networks and polynomial chaos \cite{Wiener.1938, Rosic.2019, Rosic.2011}. So far, however, there has been little discussion about the history dependency of the material state in the overall process of meta-model designing. 

To describe the complex relationships that characterize both large and small nonlinear strain behavior, the preferred surrogate model has to be capable of maintaining the past information while simultaneously approximating quantity of interest. In a nutshell, this can be achieved with the help of LSTM networks \cite{vandeWeg.2021, Zhang.2020b, Li.2020, Peng.2021}. The basic LSTM structure is based on short-term memory processes that are used to create longer-term memory and hence these networks can carry past information into the future predictions. Unlike classical feed-forward networks \cite{Schmidhuber.2015}, LSTM is providing feedback connections and is capable of maintaining the past information in the so-called memory cells \cite{Hochreiter.1997}. However, next to the varying number of LSTM cells, their architecture is rather fixed. This may often lead to over-parametrizaton, yielding that one is fitting a richer model than necessary. Consequently, from the data perspective the training of LSTM networks can be expensive and hence may lead to the over-fitting problem in the presence of data noise. 

In order to define a neural network with a flexible architecture, the network must be described by and restricted to its required complexity only. An objective of this study is therefore to investigate internal flexibility of LSTM architecture from a Bayesian perspective. In contrast to earlier works that combine Bayesian theory and recurrent neural networks from an optimization point of view \cite{He.2017, Fortunato.2017, Zhang.2020, MacKay.1992b, Doerr.2018, Nikolaev.2005, Chen.2018, Mirikitani.2009, Chatzis.2015, Gulshad.2017}, in this paper we employ Bayesian theory to explore the importance of neural connections rather than to quantify the output distribution. With the help of Bayesian sparse priors we develop a training algorithm based on the Automatic Relevance Determination (ARD) scheme \cite{Tipping.2001, QuinoneroCandela.2002, Nikolaev.2005, Liu.2008} that automatically  recognizes relevant neural network weights, and hence corresponding neural connections. The algorithm can therefore be seen as an automatic dropout program in which the optimal network parameterization is achieved. As a direct consequence, the most optimal LSTM architecture can be found, while minimizing the required training data set. The potential advantage lies in the ability to recognize model complexity and adapt the network width accordingly, reducing the risk for any over-fitting, as found in deterministic LSTM approaches.

The paper is organized as follows: In the first section the theory on LSTM networks is set out, followed by an introduction to the automatic relevance determination scheme. Subsequently, this scheme is introduced to the LSTM cell architecture. The newly defined ARD-LSTM is then applied in a structural application, after which its predictability is assessed, leading to a final conclusion on the applicability of ARD-LSTM frameworks. 

\section{Long short-term memory cell}
Let be given a physical system modeled by an equilibrium equation:
\begin{equation}
\label{abs_main_eq}
A(q,u(t))=\mathcal{F}(t), \quad u(0)=u_0
\end{equation}
in a quasi-static condition. Here, $u \in \mathcal{U}$ describes the state of the system lying in a Hilbert space $\mathcal{U}$ (for the sake of simplicity), 
$A$ is a -possibly non-linear- operator modeling the physics of the system, and $\mathcal{F}\in \mathcal{U}^*$ is some external influence (action / excitation / loading). Furthermore, we assume that the model depends on the parameter set $q \in \mathcal{Q}$  which is uncertain. The parameter set $q$ is modeled as a random vector with finite second order moments on a probability space $(\varOmega, \mathfrak{F}, \mathbb{P})$ where $\varOmega$ denotes the set of elementary events, $\mathfrak{F}$ is a $\sigma$-algebra of events, and $\mathbb{P}$ is a probability measure. With the spatial domain $\spatialdomain$ one may write:
\begin{equation}
q(\omega):\spatialdomain\times\samplespace\rightarrow\mathcal{Q}.
\end{equation}
Note that in the previous equation the external influence $\mathcal{F}$, as well as initial conditions $u_0$ can be also included in the parameter set. The theory presented further does not depend on this choice, and is general enough to cover all of mentioned cases.
 
Given $q$ and \refeq{abs_main_eq}, our interest lies in the estimation of the quantity of interest (QoI) $y(t) \in \mathcal{Y} \times \mathcal{T}$ with values $y(t)=Y(u(t),q,t)$ described by a possibly nonlinear operator $Y$. Moreover, we search for a nonlinear continuous function 
\begin{equation}
\label{eq:map_to_be_approximated}
y(t)=Y(u(t),q,t)=:G(q,t)
\end{equation}
that describes the time evolution of QoI. After the appropriate time discretization, the previous equation can be susbstituted by its incremental version
 \begin{equation}
 \label{eq:map_to_be_approximated1}
y(t_i)=G(q,t_i), \quad i=1,...,m
\end{equation}
in which $n_m$ time increments are not necessarily taken as equidistant. Due to spatial and stochastic dependence, the QoI is further discretized in both spatial and parametric space (e.g.~by the finite element method and Latin Hypercube sampling \cite{Mckay.1979}). This then leads to complete discretization 
of \refeq{eq:map_to_be_approximated}, i.e.

\begin{equation}
\forall i=1,...,n_b, \quad j=1,..,m, : \quad {\vek{y}}(\vek{q}_i,t_j) \in \mathcal{Y}_{N}, 
\end{equation}
in which $\mathcal{Y}_{N}$ denotes the discretization of the spatial domain by $N$ basis functions. Hence,
\begin{equation}
\forall j=1,..,m : \quad {\vek{y}}_{j}:={\vek{y}}(\vek{q}_i,t_j)  \quad i=1,...,n_b,
\end{equation}
is of dimension $n_b \times N$. 

Given data set $(t_i,\vek{y}_i)_{i=1}^m$ and transition parameters $\vek{\theta} \in \mathcal{O}$ (weights) we approximate \refeq{eq:map_to_be_approximated} by a recurrent neural network described by a generic layerwise function 
\begin{equation}\label{transition}
\vek{h}_i=\sigma_h(\vek{h}_{i-1}, \vek{x}_i,\vek{\theta}), \quad \sigma_h:\quad \mathcal{H} \times \mathcal{X} \times \mathcal{O}\mapsto \mathcal{H}
\end{equation}
and the prediction function  
\begin{equation}\label{prediction}
\hat{\vek{y}}_i = \sigma_y(\vek{h}_i,\vek{\xi}), \quad \sigma_y: \quad \mathcal{H}\times \mathcal{O} \mapsto \mathcal{Y}_{n_f},
\end{equation}
with $\vek{\xi}$ being the prediction parameters (weights). Here, $\vek{h}_i \in \mathcal{H}$ is of dimension $n_b \times n_m$ ($n_b$ is the batch size and $n_m$ is the dimension of state or also known as the number of units) denotes the so-called hidden state at time $t_i$ with $\vek{h}_0 \in \mathcal{H}$ being the initial hidden state, which can either be learned as a parameter or initialized to some fixed vector. Input $\vek{x}_i \in \mathcal{X}$ is of the dimension $n_b\times n_f$ ( $n_f$ being the number of features) are the inputs at time $t_i$. Furthermore, ($\vek{\theta},\vek{\xi}$) are intra-dependent as further explained in the text.

Following previous statements, the generic approximation of \refeq{eq:map_to_be_approximated} can be formulated as a layerwise composition of functions
\begin{equation}\label{prediction1}
\hat{\vek{y}}_i= \sigma_y \circ \sigma_h^i \circ ... \circ \sigma_h^1 
\end{equation}
with $\sigma_h^i: \textrm{ }\mathcal{H} \times \mathcal{X} \times \mathcal{O}\rightarrow \mathcal{H}$ being the transition function at time $t_i$ for all $i=1,..,m$ besides $i=1$ in which $\sigma_h^1$ denotes the identity map. In this formulation the unknown variable is the network parameter set. Its estimation is usually done in a classical mean squared sense by minimizing the loss function
\begin{equation}\label{mse}
\begin{aligned}
\mathcal{J} &= \sum_{i=1}^m J(\vek{y}_i,\hat{\vek{y}}_i),\\
J(\vek{y}_i,\hat{\vek{y}}_i)&=\frac{1}{2} \langle \vek{y}_i-\hat{\vek{y}}_i(\vek{\theta}_i,\vek{\xi}_i),\vek{y}_i-\hat{\vek{y}}_i(\vek{\theta}_i,\vek{\xi}_i)\rangle.
\end{aligned}
\end{equation}
The minimization is usually performed by use of gradient-like approaches also known as backpropagation \cite{Williams.1995b, Werbos.1988}. One major drawback of this training approach is that the error exponentially decays/grows with time. Hence, the memory effect in the output resembled in the recurrent relation in \refeq{prediction1} is of the short-term type \cite{Hochreiter.1997}. 

\begin{figure}[ht!]
\centering
\includegraphics[width=0.95\textwidth]{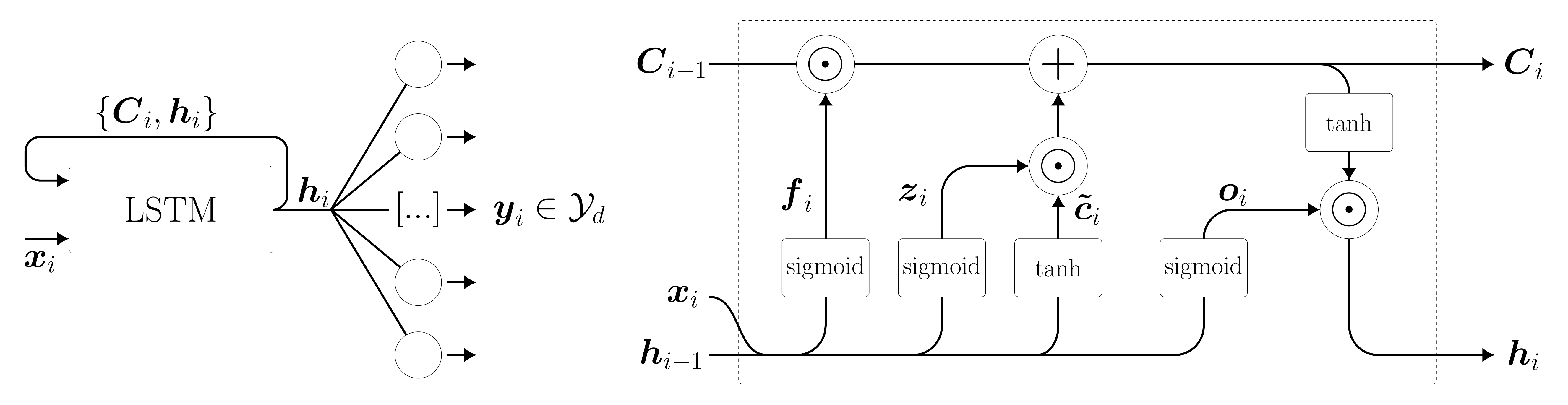}
\caption{LSTM-based neural network (left) and the LSTM cell (right), comprising the continuous cell state $\vek{c}_i$ and hidden state $\vek{c}_i$ for time $t_i$, adapted from \cite{vandeWeg.2021}.}
\label{fig:LSTMcell}
\end{figure}

To overcome the previously mentioned issue,  a series of gating mechanisms for hidden state evolution is added to the overall structure in \refeq{transition}-\refeq{prediction} to control the error, see \cite{Hochreiter.1997}. This is achieved by substituting the transition function presented in \refeq{transition}, which defines the LSTM cell with (see \reffig{fig:LSTMcell}):
\begin{equation}
\begin{aligned}
\label{lstm_transition}
\vek{f}_{i}&=\sigma _{f}(\boldsymbol{\mathit{\Phi}}_{i} \vek{w}_{f} +\vek{b}_f) \\
 \vek{z}_{i}&=\sigma _{i}(\boldsymbol{\mathit{\Phi}}_{i}  \vek{w}_{z} +\vek{b}_z) \\
 {\tilde{\vek{c}}}_{i}&=\sigma_{\tilde{c}}(\boldsymbol{\mathit{\Phi}}_{i} \vek{w}_{\tilde{c}}+\vek{b}_c  ) \\
  \vek{o}_{i}&= \sigma_{o}( \boldsymbol{\mathit{\Phi}}_{i}  \vek{w}_o+\vek{b}_o) \\
 \vek{C}_{i}&=\vek{f}_{i}\odot \vek{c}_{i-1}+\vek{z}_{i}\odot {\tilde{\vek{c}}}_{i}\\
  \vek{h}_{i}&= \vek{o}_{i}\odot \sigma_{c}(\vek{c}_{i})
\end{aligned}
\end{equation}
in which  $\boldsymbol{\mathit{\Phi}}_{i} = \left[ \vek{x}_{i}, \vek{h}_{i-1}\right]$.
Here, $\odot$ is the Hadamard product, $\sigma_{\mathrm{g}}$ is the the activation function corresponding to the gate $\mathrm{g} \in \left\lbrace \mathrm{f}, \mathrm{z}, \mathrm{\tilde{c}}, \mathrm{o} \right\rbrace $, $\vek{x}_i$ is the input state and $\vek{h}_i$ is the hidden state at time $t_i$. We may also distinguish the forget gate's activation vector $\vek{f}_{i}$, the input gate's activation vector $\vek{z}_{i}$, the output gate's activation vector $\vek{o}_{i}$, the cell candidate activation vector $\tilde{\vek{c}}_i$, and the cell state $\vek{c}_i$ \cite{Gers.2000}.
The prediction equation for the output state $\vek{y}_i$ stays the same as in \refeq{prediction} and hence  the layerwise prediction in \refeq{prediction1} for an LSTM-based neural network rewrites to
 \begin{equation}\label{predictionLSTM}
\hat{\vek{y}}_i= \sigma_y \circ \sigma_{\hbar}^i \circ ... \circ \sigma_{\hbar}^1 
\end{equation}
in which $\sigma_{\hbar}^i$ describes the transition function implicitly defined by \refeq{lstm_transition} at the time $t_i, \textrm{ }i=1,...,m$. 

The unknown weights can be estimated using a gradient-based iteration:
\begin{equation}
\label{weightUpdate}
\vek{w}^{new}_\mathrm{g} = \vek{w}^{old}_\mathrm{g} - \lambda\frac{\partial \mathcal{J}}{\partial \vek{w}^{old}_\mathrm{g}}, \quad \mathrm{g} \in \left\lbrace \mathrm{f}, \mathrm{z}, \mathrm{\tilde{c}}, \mathrm{o} \right\rbrace.
\end{equation}
Here, $\lambda$ is the learning rate and
\begin{equation}
\frac{\partial \mathcal{J}}{\partial \vek{w}_\mathrm{g}} = \sum_i \boldsymbol{\mathit{\Phi}}_{i}^T \odot \partial \vek{g}_{i},\quad \vek{g} \in \left\lbrace \vek{f}, \vek{z}, \tilde{\vek{c}}, \vek{o} \right\rbrace
\end{equation}
where $\vek{g}_{i}$ generically denotes the output of the corresponding gate in LSTM cell, see \refeq{lstm_transition}. Gradients $\partial \vek{g}_i$ can be obtained by use of the backpropagation algorithm:
\begin{equation}
\begin{aligned}
\label{eq:LSTM_backprop}
\partial \vek{h}_i &= \frac{\partial \boldsymbol{J}_i}{\partial \vek{h}_i} + \nabla \vek{h}_i \\
\partial \vek{C}_i &= \partial \vek{h}_i \odot \vek{o}_i \odot (1-\mathrm{tanh}^2(\vek{C}_i))+\partial \vek{C}_{i+1} \odot \vek{f}_{i+1} \\
\partial \tilde{\vek{c}}_i &= \partial \vek{C}_i \odot \vek{z}_i \odot (1-\tilde{\vek{c}}_i^{2})\\
\partial \vek{z}_i &= \partial \vek{C}_i \odot \tilde{\vek{c}}_i \odot \vek{z}_i \odot (1-\vek{z}_i) \\
\partial \vek{f}_i &= \partial \vek{C}_i \odot \vek{C}_{i-1} \odot  \vek{f}_i \odot (1-\vek{f}_i) \\
\partial \vek{o}_i &= \partial \vek{h}_i \odot \mathrm{tanh}(\vek{C}_i) \odot \vek{o}_i \odot (1-\vek{o}_i).
\end{aligned}
\end{equation}
Here, $\frac{\partial \boldsymbol{J}_i}{\partial \vek{h}_t}$ is the  predictive gradient, whereas $\nabla \vek{h}_i$ is the recurrent gradient from the subsequent time step, described by:
\begin{equation}
\nabla \vek{h}_{i} = \sum_{\mathrm{g}} \vek{r}_{\mathrm{g}} \partial \vek{g}_{i},
\end{equation}
with $\vek{r}_{\mathrm{g}} := \left[\vek{w}_{\mathrm{g},n_f},\vek{w}_{\mathrm{g},n_f+1},...,\vek{w}_{\mathrm{g},n_f+n_m} \right]^T$. The classical backpropagation -as described in \\ \refeq{eq:LSTM_backprop}- results in a point estimate for weights. These point estimates describe neural connections of a proxy model, optimized to and hence valid for the available data set only \cite{Lehmann.1998}. To properly approximate any physical system, especially in complex finite element applications, a large data set must often be considered. A smaller data set increases the uncertainty on the choice and design of a proxy model. By implying point estimate driven approach, one endeavors to ignore the predictive uncertainty by a fixed number of parameters that describe a pre-defined model complexity. As a result, a point estimate driven neural network will, in any application, enforce a high risk for either under- or over-fitting. 

Dropout, or methods alike, impose a randomized deletion of weights in optimization \cite{Srivastava.2014}. Weight co-adaptation is thereby prevented. Dropout, although often helpful but mathematically less supported, does not provide post training width reduction and thus generalization capability that one desires from a neural network. It therefore fails to adapt the model complexity and architecture to the problem description. To provide the required network architecture flexibility and auto-adaption capability, one needs to evaluate the relevance of every individual weight (connection) during optimization. Therefore, one may follow a Bayesian approach to obtain a scheme that automatically determines the relevance of each weight in a neural network.

\section{Automatic relevance determination}
\begin{figure}[ht!]
\centering
\includegraphics[width=0.7\textwidth]{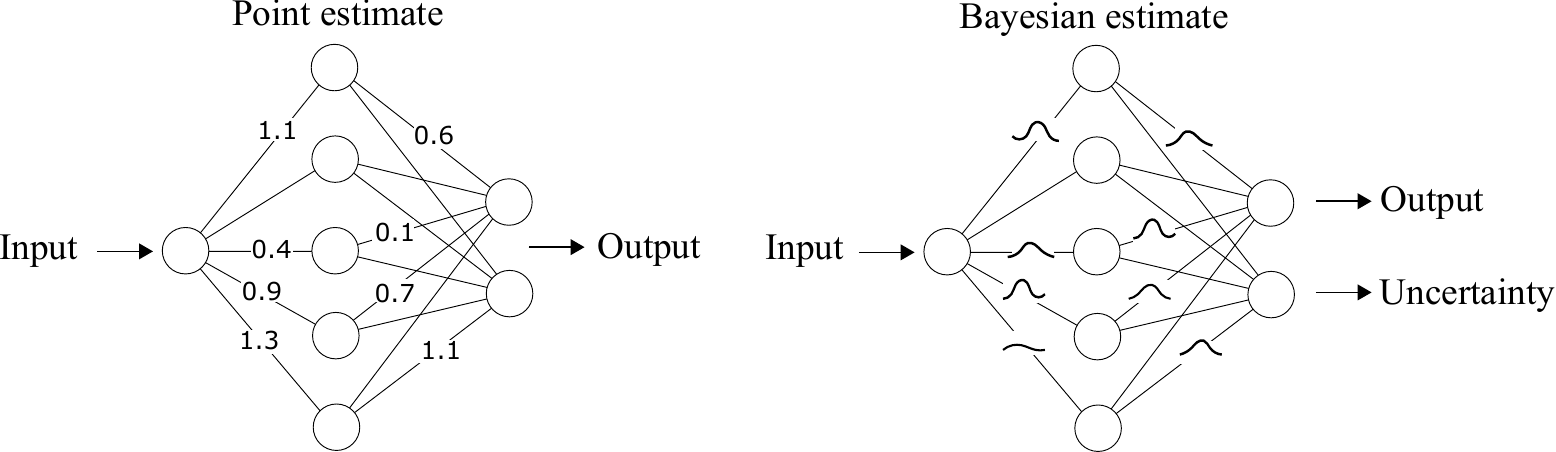}
\caption{Comparison between a plain feed-forward neural network with point estimate weights (left) and probabilistic weights (right), adapted from \cite{Blundell.2015}.}
\label{fig:prob}
\end{figure}

The first step towards a flexible network architecture is to describe the loss function (see \refeq{mse}) in a more generalized, probabilistic sense. As the weights $\vek{w}$ in a general LSTM are unknown, we may assume them as uncertain, and model them as random variables $\vek{w}(\omega_w)$ (i.e.~all collected unknown weight matrices and biases) with finite variance on a probability space $(\varOmega_w,\mathfrak{F}_w,\mathbb{P}_w)$, see \reffig{fig:prob}. 

By argument $\omega_w$ we denote the uncertainty propagation of weights $\vek{w}(\omega_w)$ through the network. With the help of \refeq{prediction1} one can predict the network output
\begin{equation}
\label{eq:prediction15}
\hat{\vek{y}}_i(\omega_w,\omega_e)={\vek{y}}_i(\omega_w)+\vek{\varepsilon}_i(\omega_e)
\end{equation}
in which $\vek{\varepsilon}_i(\omega_e)$ denotes the prediction of the measurement/modeling error at time $t_i$ and is described as a random variable on a probability space $(\varOmega_e,\mathfrak{F}_e,\mathbb{P}_e)$. Denoting the joint space $(\varOmega_s, \mathfrak{F}_s,\mathbb{P}_s)$ with $\varOmega_s:=\varOmega_w\times \varOmega_e$, one may rewrite \refeq{eq:prediction15} as:
\begin{equation}\label{final_prediction}
\hat{\vek{y}}_i(\omega_s)=\vek{y}_i(\omega_s)+\vek{\varepsilon}_i(\omega_s).
\end{equation}
with $\vek{\varepsilon}_i(\omega_s)$ being independent of $\vek{\varepsilon}_j(\omega_s)$ for $i\neq j$, and $\vek{w}(\omega_s)$ now being represented on a joint space. Note that in this manner we account for the data noise in the weight estimation.  

The weights can be assimilated with the data set by use of Bayes theorem:
\begin{equation}
\label{eq:bayesfunc}
p(\vek{w}\vert \vek{y}_1,....,\vek{y_m} )=\frac{p(\vek{y}_1,....,\vek{y_m} \vert \vek{w})p(\vek{w})}{P(\vek{y}_1,....,\vek{y_m} )},
\end{equation}
in which the uncertain weights are estimated, i.e.~updated. In other words, the prior probability density function $p(\vek{w})$ is updated to the posterior probability density function $p(\vek{w}\vert \vek{y}_1,....,\vek{y_m} )$ given the likelihood function $p(\vek{y}_1,....,\vek{y_m} \vert \vek{w})$ and the evidence \\$P(\vek{y}_1,....,\vek{y_m} )$ (i.e.~the normalization factor). The likelihood function $p(\vek{y}_1,....,\vek{y_m} \vert \vek{w})$ describes how likely is the data set $\vek{y}_1,....,\vek{y_m} $ given the parameters $\vek{w}$, and is a function defined by a shape of the measurement/modelling error.  In a special case when the likelihood function is described by a zero-mean Gaussian with identity covariance $C_\varepsilon=I$, the maximum likelihood loss function matches the optimization function given in \refeq{mse}. Furthermore, if 
$p(\vek{w})$ follows the normal distribution $p(\vek{w})\sim \mathcal{N}(\hat{\vek{w}},C_w)$, and the likelihood is zero-mean Gaussian with the covariance $C_\varepsilon$, the maximum a posteriori estimate (MAP) represents the regularized form of \refeq{mse}
\begin{equation}\label{mseLSTM1}
\mathcal{J}(\omega_s)+\frac{1}{2} \langle \vek{w}-\hat{\vek{w}},\vek{w}-\hat{\vek{w}}\rangle_{{C}_w}
\end{equation}
in which $\langle x,z\rangle_C:= x^*C^{-1} z$.
 With the help of the previously defined loss function one can take into account the measurement noise, as well as the prior knowledge on $\vek{w}$ into the neural network formulation. However, the last term in \refeq{mseLSTM1} has a role of $\ell_2$ regularisation, meaning that the posterior estimate is heavily constrained on the Gaussian space of prior random variables, and is not sparse. Hence, to estimate the posterior one requires as many data points as unknown coordinates in $\vek{w}$. To introduce the sparsity to the solution we use the automatic relevance determination (ARD) scheme as described in \cite{Tipping.2001}.

The ARD scheme is designed for linear estimation problems in which the measurements (data) are linear in the parameter set. Following \refeq{predictionLSTM} one may conclude that this is not true in the LSTM case due to nonlinearity of the activation functions. Translation of the LSTM cell gates to an automatic relevance determination scheme can be done by defining local target functions using \refeq{lstm_transition}:
\begin{equation}
\label{eq:ARDLSTM}
\vek{s}_{\mathrm{g}i} := \boldsymbol{\mathit{\Phi}}_{i} \vek{w}_{\mathrm{g}}+\vek{b}_{\mathrm{g}}, \quad \vek{w}_{\mathrm{g}i}:=[\vek{w}_{\mathrm{g}}, \vek{b}_{\mathrm{g}}], \quad g  \in \left\lbrace {f}, {z}, \tilde{{c}}, {o} \right\rbrace
\end{equation}
with the local target function $\vek{s}_{\mathrm{g}i}$.
Accordingly, the gate output can be defined by:
\begin{equation}\label{noneq}
\vek{g}_{i} := \sigma_\mathrm{g}(\vek{s}_{\mathrm{g}i}).
\end{equation}
Thus by linearizing $\vek{g}_{i}$ we search for the weight $\vek{w}_{\mathrm{g}i}$ in an ARD manner. However, the main problem is that the target data $\vek{s}_{\mathrm{g}i}$ is unknown a priori. Therefore, $\vek{s}_{\mathrm{g}i}$ has to be obtained by solving \refeq{noneq} as described later in the text. Due to simplicity in notation in the further text we will use $\mathit{\Phi}_{i} \vek{w}_{\mathrm{g}i}$ to denote the sum $\mathit{\Phi}_{i} \vek{w}_{\mathrm{g}}+\vek{b}_{\mathrm{g}}$.
 
To impose sparsity of the weights in \refeq{eq:ARDLSTM}, we introduce $\ell_1$ regularization instead of $\ell_2$ by taking the Laplace prior \cite{Kaban.2007}. A Laplace distribution with mean $\nu$ and variance $\varsigma>0$ takes the form:
\begin{equation}
\label{eq:laplace_basis}
\eta (\omega_w) \sim \frac{1}{2\varsigma} e^{-\frac{\vert \omega_w-\nu \vert}{\varsigma}},
\end{equation}
and is imposed on each of the element of the weight matrices $\vek{w}_{\mathrm{g}i}$ in \refeq{eq:ARDLSTM} (in total $gn_m(n_f+n_m)+n_m)$ unknown independent weights for the network). After substituting of \refeq{eq:laplace_basis} in \refeq{eq:ARDLSTM} and applying the Bayes rule, for a zero-mean prior one obtains the logarithm of the posterior in a form:
\begin{equation}
\label{eq:log_MAP}
\mathrm{log} p(\vek{w}_{\mathrm{g}i} \vert \vek{s}_{\mathrm{g}i}) = -\sigma_{\mathrm{g}i}^{-1}\langle \vek{s}_{\mathrm{g}i} - \boldsymbol{\mathit{\Phi}}_{i}\vek{w}_{\mathrm{g}}-\vek{b}_{\mathrm{g}}, \vek{s}_{\mathrm{g}i} -  \boldsymbol{\mathit{\Phi}}_{i}\vek{w}_{\mathrm{g}}-\vek{b}_{\mathrm{g}} \rangle - \frac{1}{b} \vert \vek{w}_{\mathrm{g}i} \vert + \mathrm{const.}
\end{equation}
However, the posterior in \refeq{eq:log_MAP} cannot be evaluated directly as a Laplace distribution is not conjugate with the Gaussian likelihood function. This can be overcome by defining hyperpriors that emulate a Laplace distribution. For weights following a Gaussian distribution $p(\vek{w}_{\mathrm{g}i}\vert \vek{\alpha}_{\mathrm{g}i}) =\mathcal{N}(0,\vek{\alpha}_{\mathrm{g}i}^{-1})$, one can assure a Laplace emulation using a variance that is described by a uniform (non-informative) type of hyperprior $p(\vek{\alpha}_{\mathrm{g}i})$. In logarithmic scale this matches with a non-informative Gamma distribution \cite{Tipping.2001}:
\begin{equation}
\label{eq:alpha_hyperprior}
p(\vek{\alpha}_{\mathrm{g}i}) = \prod_{k} \prod_{l} \Gamma(a)^{-1}b^a\alpha_{gikl}^{a-1} \mathrm{exp}(-b \alpha_{gikl}) 
\end{equation}
in which $\Gamma(a)= \int_0^{\infty} t^{a-1}\ \mathrm{exp}(-t)\mathrm{d}t$ and $a$, $b$ are the shape and scale parameters, respectively. 

Assuming that \refeq{eq:ARDLSTM} is described by noise (due to modelling or data error), we may add zero-mean Gaussian additive term with the variance $\vek{var}_{gi}$ that is independent for each gate and each time step. As the noise variance  $ \vek{var}_{gi}$ is unknown, one can model its inverse $\vek{\beta}_{gi} := \vek{var}_{gi}^{-1}$ by a hyperprior $p(\vek{\beta}_{\mathrm{g}i})$, similarly defined as $\vek{\alpha}_{gi}$:
\begin{equation}
\label{eq:beta_hyperprior}
p(\vek{\beta}_{\mathrm{g}i}) = \prod_{l=0} \Gamma(c)^{-1}d^c\beta_{gil}^{c-1} \mathrm{exp}(-d \textrm{ }\beta_{gil}),
\end{equation}
with $c$, $d$ being respectively the shape and scale parameter. With $a=b=c=d=0$ one obtains a uniform hyperprior.

\begin{figure}[ht!]
\centering
\begin{subfigure}[c]{0.3\textwidth}
\centering
\includegraphics[width=\textwidth]{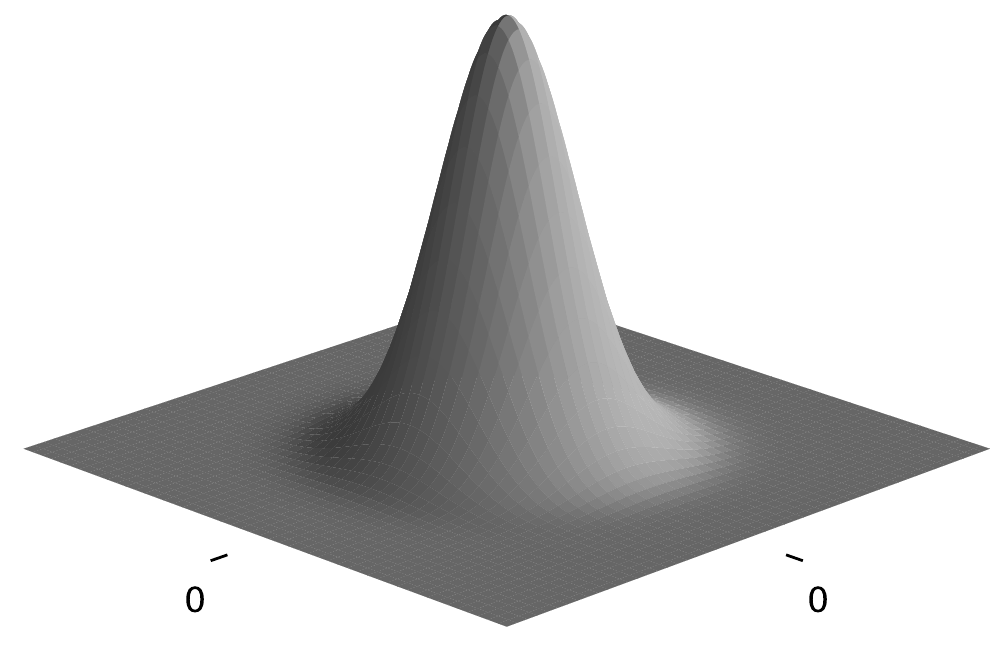}
\subcaption{Multivariate Gaussian.}
\label{fig:Multivariate_Gaussian}
\end{subfigure}
\begin{subfigure}[c]{0.3\textwidth}
\centering
\includegraphics[width=\textwidth]{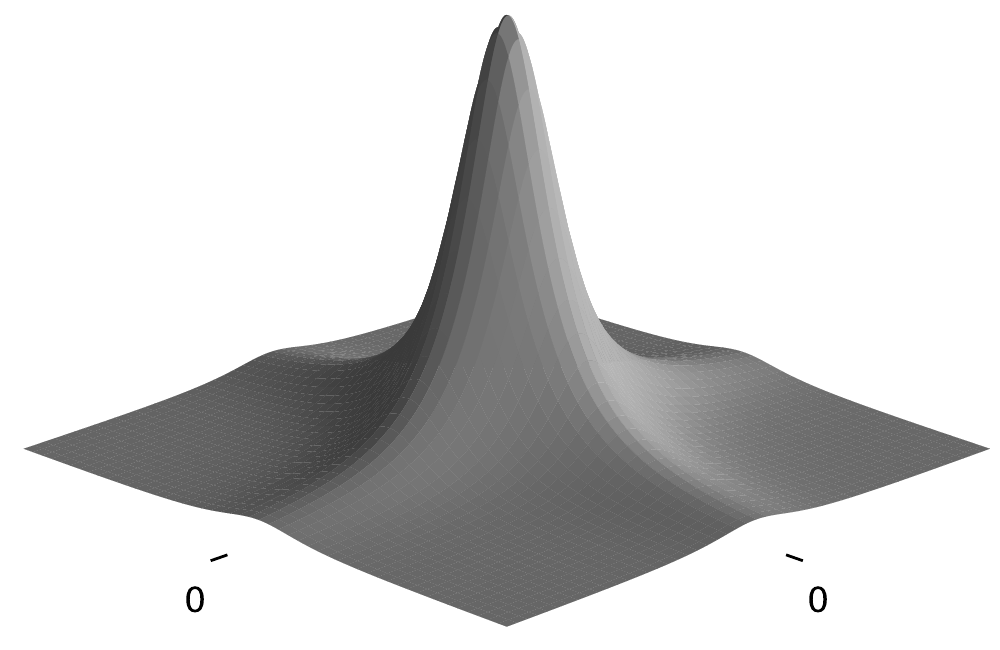}
\subcaption{Multivariate Student-t.}
\label{fig:Multivariate_Student_t}
\end{subfigure}
\begin{subfigure}[c]{0.3\textwidth}
\centering
\includegraphics[width=\textwidth]{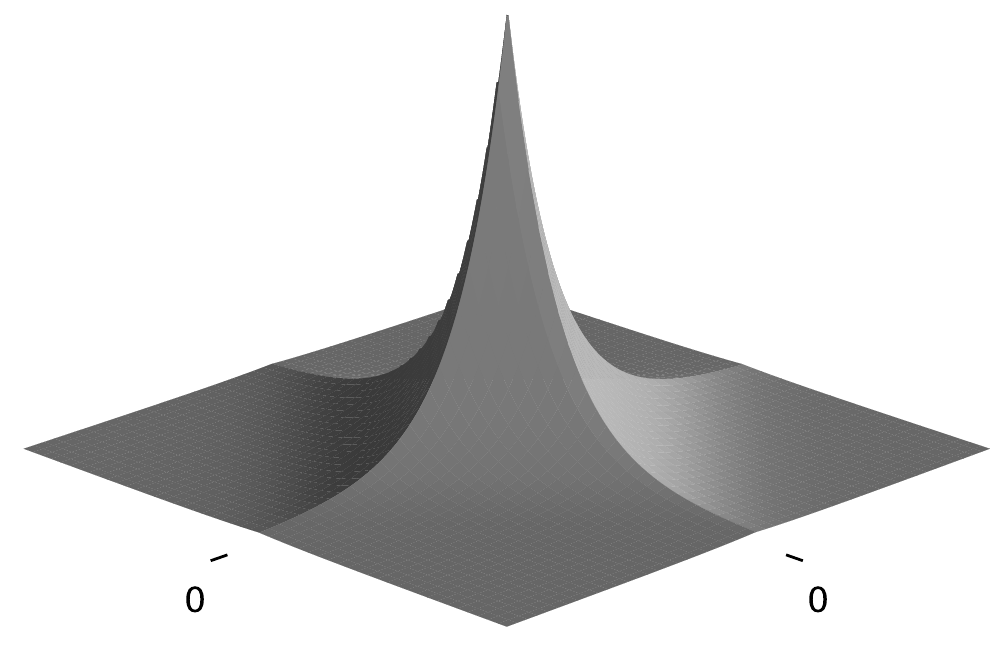}
\subcaption{Multivariate Laplace.}
\label{fig:Multivariate_Laplace}
\end{subfigure}
\caption{Multivariate prior distributions, with (\protect\subref{fig:Multivariate_Gaussian}) a multivariate Gaussian prior $p(\vek{w}\vert \vek{\alpha})$; (\protect\subref{fig:Multivariate_Student_t}) a Student-t prior $p(\vek{w})$ and (\protect\subref{fig:Multivariate_Laplace}) a heavily sparsity favoring Laplace distribution.}
\label{fig:Multivariate_distributions}
\end{figure}

The Bayesian description over all unknowns then reads:
\begin{equation}
\label{eq:hierarchicalbayes}
p(\vek{w}_{\mathrm{g}i},\vek{\alpha}_{\mathrm{g}i},\vek{\beta}_{\mathrm{g}i} \vert \vek{s}_{\mathrm{g}i}) = \frac{p(\vek{s_{\mathrm{g}i}} \vert\vek{w}_{\mathrm{g}i},\vek{\alpha}_{\mathrm{g}i},\vek{\beta}_{\mathrm{g}i}) p(\vek{w}_{\mathrm{g}i},\vek{\alpha}_{\mathrm{g}i},\vek{\beta}_{\mathrm{g}i})}{P(\vek{s}_{\mathrm{g}i})}.
\end{equation}
From this, the posterior distribution can be modeled as a product of the posterior over the weights $p(\vek{w}_{\mathrm{g}i} \vert \vek{s}_{\mathrm{g}i}, \vek{\alpha}_{\mathrm{g}i},\vek{\beta}_{\mathrm{g}i})$ and the posterior over the hyperparameters $p(\vek{\alpha}_{\mathrm{g}i},\vek{\beta}_{\mathrm{g}i} \vert \vek{s}_{\mathrm{g}i})$:
\begin{equation}
\label{eq:postdecompose}
p(\vek{w}_{\mathrm{g}i},\vek{\alpha}_{\mathrm{g}i},\vek{\beta}_{\mathrm{g}i} \vert \vek{s}_{\mathrm{g}i}) = p(\vek{w}_{\mathrm{g}i} \vert \vek{s}_{\mathrm{g}i}, \vek{\alpha}_{\mathrm{g}i},\vek{\beta}_{\mathrm{g}i})p(\vek{\alpha}_{\mathrm{g}i},\vek{\beta}_{\mathrm{g}i} \vert \vek{s}_{\mathrm{g}i}).
\end{equation}
The first term in \refeq{eq:postdecompose} is obtained via:
\begin{equation}
\label{eq:post_weights}
p(\vek{w}_{\mathrm{g}i} \vert \vek{s}_{\mathrm{g}i}, \vek{\alpha}_{\mathrm{g}i},\vek{\beta}_{\mathrm{g}i}) = \frac{p(\vek{s}_{\mathrm{g}i} \vert \vek{w}_{\mathrm{g}i},\vek{\beta}_{\mathrm{g}i}) p(\vek{w}_{\mathrm{g}i} \vert \vek{\alpha}_{\mathrm{g}i})}{p(\vek{s}_{\mathrm{g}i} \vert \vek{\alpha}_{\mathrm{g}i},\vek{\beta}_{\mathrm{g}i})},
\end{equation}
which presents a convolution of Gaussians that can be computed directly. As a final outcome we obtain a multivariate Gaussian distribution over the weights with the covariance and mean respectively given by
\begin{equation}
\label{eq:musigma}
\begin{aligned}
\vek{\Sigma}_{\mathrm{g}i} &:= (\vek{\beta}_{\mathrm{g}i} \odot \boldsymbol{\mathit{\Phi}}_{i}^T \boldsymbol{\mathit{\Phi}}_{i} + \vek{\alpha}_{\mathrm{g}i}\vek{I})^{-1} \\
\vek{\mu}_{\mathrm{g}i} &:= \vek{\beta}_{\mathrm{g}i} \odot \vek{\Sigma}_{\mathrm{g}i} \boldsymbol{\mathit{\Phi}}_{i}^T \vek{s}_{\mathrm{g}i}. \\
\end{aligned}
\end{equation}
To evaluate the effect of hyperprior $p(\vek{\alpha}_{\mathrm{g}i})$ on Gaussian prior $p(\vek{w}_{\mathrm{g}i} \vert \vek{\alpha}_{\mathrm{g}i})$ (see \reffig{fig:Multivariate_Gaussian}) one needs to integrate out $\vek{\alpha}_{\mathrm{g}i}$:
\begin{equation}
\begin{aligned}
p(\vek{w}_{\mathrm{g}i}) &= \int p(\vek{w}_{\mathrm{g}i} \vert \vek{\alpha}_{\mathrm{g}i})p(\vek{\alpha}_{\mathrm{g}i})\mathrm{d} \vek{\alpha}_{\mathrm{g}i} \\
&= \frac{\vek{b}_{\mathrm{g}i}^{\vek{a}_{\mathrm{g}i}} \Gamma(\vek{a}_{\mathrm{g}i} + \frac{1}{2})}{(2\pi)^{\frac{1}{2}} \Gamma(\vek{a}_{\mathrm{g}i})} (\vek{b}_{\mathrm{g}i} + \frac{ \vek{w}_{\mathrm{g}i}^2}{2})^{-(\vek{a}_{\mathrm{g}i}+\frac{1}{2})},
\end{aligned}
\end{equation}
which results in a Student-t distribution over $p(\vek{w}_{\mathrm{g}i})$ for $\vek{a}_{\mathrm{g}i}, \vek{b}_{\mathrm{g}i} \neq 0$, see \reffig{fig:Multivariate_Student_t}. For $\vek{a}_{\mathrm{g}i} = \vek{b}_{\mathrm{g}i} = 0$ one obtains $p(\vek{w}_{\mathrm{g}i}) \propto 1 / \sqrt{\vek{w}_{\mathrm{g}i}^2}$, which strongly favors sparsity and shows a high similarity with the Laplace prior, see \reffig{fig:Multivariate_Laplace} \cite{Williams.1995, Gerven.2009}.

Given the first term in \refeq{eq:postdecompose}, it remains to estimate $p(\vek{\alpha}_{\mathrm{g}i},\vek{\beta}_{\mathrm{g}i} \vert \vek{s}_{\mathrm{g}i})$. Assuming independence of $p(\vek{\alpha}_{\mathrm{g}i})$ and $p(\vek{\beta}_{\mathrm{g}i})$, one may further write:
\begin{equation}
\label{eq:opti_obj}
p(\vek{\alpha}_{\mathrm{g}i},\vek{\beta}_{\mathrm{g}i} \vert \vek{s}_{\mathrm{g}i}) \propto p(\vek{s}_{\mathrm{g}i} \vert \vek{\alpha}_{\mathrm{g}i},\vek{\beta}_{\mathrm{g}i}) p(\vek{\alpha}_{\mathrm{g}i}) p(\vek{\beta}_{\mathrm{g}i}).
\end{equation}
For computational simplicity one may approximate the posterior by the most probable value, according to \cite{Tipping.2001, Rosic.2019}:
\begin{equation}
p(\vek{\alpha}_{\mathrm{g}i},\vek{\beta}_{\mathrm{g}i} \vert \vek{s}_{\mathrm{g}i}) \approx \delta(\vek{\alpha}_{_{\mathrm{g}i},\mathrm{MP}},{\vek{\beta}}_{_{\mathrm{g}i},\mathrm{MP}}).
\end{equation}
which then reads:
\begin{equation}
\delta(\vek{\alpha}_{\mathrm{g}i},\vek{\beta}_{\mathrm{g}i} \vert \vek{s}_{\mathrm{g}i})_{\mathrm{max}} = \underset{\alpha, \beta}{\operatorname{argmax}}\ \mathrm{log}p(\vek{s}_{\mathrm{g}i} \vert \mathrm{log}\vek{\alpha}_{\mathrm{g}i},\mathrm{log}\vek{\beta}_{\mathrm{g}i}).
\end{equation}
This completes the description of the decomposed posterior in  \refeq{eq:postdecompose}. The objective is to maximize the marginal likelihood $p(\vek{s}_{\mathrm{g}i} \vert \vek{\alpha}_{\mathrm{g}i},\vek{\beta}_{\mathrm{g}i}) = \int p(\vek{s}_{\mathrm{g}i} \vert \vek{w}_{\mathrm{g}i}, \vek{\beta})p(\vek{w}_{\mathrm{g}i} \vert \vek{\alpha}_{\mathrm{g}i})\mathrm{d}\vek{w}_{\mathrm{g}i}$. After taking the logarithm one finds the objective function \cite{Tipping.2001}:
\begin{equation}
\label{eq:marginal_likelihood_cost}
\mathcal{L}_{\mathrm{g}i} = -\frac{1}{2}\left\lbrace \mathrm{log} \vert \beta_{\mathrm{g}i}^{-1}\vek{I} + \boldsymbol{\mathit{\Phi}}_{i}(\vek{\alpha}_{\mathrm{g}i}\vek{I})^{-1}\boldsymbol{\mathit{\Phi}}_{i}^T \vert + \vek{s}_{\mathrm{g}i}^T (\vek{\beta}_{\mathrm{g}i}^{-1}\vek{I} + \boldsymbol{\mathit{\Phi}}_{i}(\vek{\alpha}_{\mathrm{g}i}\vek{I})^{-1}\boldsymbol{\mathit{\Phi}}_{i}^T)^{-1}\vek{s}_{\mathrm{g}i}\right\rbrace.
\end{equation}

Finally, $\vek{\alpha}_{\mathrm{g}i}$ and $\vek{\beta}_{\mathrm{g}i}$ can be estimated via gradient based optimization to obtain \cite{Tipping.2001}:
\begin{equation}
\label{eq:alphacompute}
\vek{\alpha}_{\mathrm{g}ik} = \frac{1}{\vek{\mu}_{\mathrm{g}ik}^2 + \vek{\Sigma}_{\mathrm{g}ikk}}
\end{equation}
and
\begin{equation}
\label{eq:variancecompute}
\vek{\beta}_{\mathrm{g}i} = \frac{D - \sum_k \vek{\gamma}_{\mathrm{g}ik}}{\Vert \vek{s}_{\mathrm{g}i} - \boldsymbol{\mathit{\Phi}}_{i}\vek{\mu}_{\mathrm{g}i}\Vert^2}.
\end{equation}
Parameter $\vek{\gamma}_{\mathrm{g}ik}$ is defined by \cite{Tipping.2001}: 
\begin{equation}
\label{eq:pruning_setup}
\vek{\gamma}_{\mathrm{g}ik} = 1 -\vek{\alpha}_{\mathrm{g}ik} \odot \Sigma_{\mathrm{g}ikk}.
\end{equation}
If $w_{\mathrm{g}ikl}$ is majorly constrained by its prior $\alpha_{\mathrm{g}ikl}$ while being uncertain in its likelihood ($\Sigma_{\mathrm{g}ikkl} \approx \alpha_{\mathrm{g}ikl}^{-1}$), $\gamma_{\mathrm{g}ikl} \rightarrow \num{0}$ and hence the corresponding weight can be set to zero.

\subsection{Posterior predictive}

With optimized most probable values for $\vek{\alpha}_{\mathrm{g}i}$ and $\vek{\beta}_{\mathrm{g}i}$ one can subsequently compute the posterior predictive $p(\hat{\vek{s}}_{\mathrm{g}i}^{*})$ using the likelihood $p(\hat{\vek{s}}_{\mathrm{g}i}^{*} \vert \vek{w}_{\mathrm{g}i}, \vek{\beta}_{_{\mathrm{g}i}\mathrm{MP}})$ and posterior over the weights $p(\vek{w}_{\mathrm{g}i} \vert \vek{s}_{\mathrm{g}i},\vek{\alpha}_{_{\mathrm{g}i}\mathrm{MP}}, \vek{\beta}_{_{\mathrm{g}i}\mathrm{MP}})$. With both the likelihood and the posterior over the weights represented by Gaussians, one can derive the posterior predictive $p(\hat{\vek{s}}_{\mathrm{g}i}^{*} \vert \vek{s}_{\mathrm{g}i}, \vek{\alpha}_{{\mathrm{g}i}\mathrm{MP}},\vek{\beta}_{{\mathrm{g}i}\mathrm{MP}})$, giving \cite{Tipping.2001}:
\begin{equation}
\label{eq:pred_overview}
p(\hat{\vek{s}}_{\mathrm{g}i}^{*} \vert \vek{s}_{\mathrm{g}i}, \vek{\alpha}_{\mathrm{g}i,\mathrm{MP}},\vek{\beta}_{\mathrm{g}i,\mathrm{MP}}) = \mathcal{N}(\hat{\vek{s}}_{\mathrm{g}i}^{*} \vert \overline{\vek{s}}_{\mathrm{g}i}^{*},\vek{\Sigma}_{\mathrm{g}i}^*),
\end{equation}
with
\begin{equation}
\label{eq:ARDpredict}
\begin{aligned}
\overline{\vek{s}}_{\mathrm{g}i}^{*} &= \boldsymbol{\mathit{\Phi}}_{i}(\vek{x_i^{*}})\vek{\mu}_{\mathrm{g}i}\\
(\vek{\Sigma}^*)_{\mathrm{g}i}^{2} &= \vek{\beta}_{\mathrm{g}i,\mathrm{MP}}^{-1} +  \boldsymbol{\mathit{\Phi}}_{i}(\vek{x_i^{*}})^T\vek{\Sigma}_{\mathrm{g}i}  \boldsymbol{\mathit{\Phi}}_{i}(\vek{x_i^{*}}),
\end{aligned}
\end{equation}
referring to $\vek{x}_i^{*}$, $\overline{\vek{s}}_{\mathrm{g}i}^{*}$ and $(\vek{\Sigma}^*)_{\mathrm{g}i}^{2} $ as the input feature vector, predictive mean and variance, respectively. As the output activation function $\sigma_y$ is purely linear, the prediction $\hat{\vek{y}}_i$ of the network is directly given for its input $\boldsymbol{\mathit{\Psi}}_{i} = [1, \vek{h}_{i,\mathrm{MP}}]$:
\begin{equation}
\label{eq:outputPredict}
\hat{\vek{y}}_i = \boldsymbol{\mathit{\Psi}}_{i}(\vek{x}_i^*) \vek{w}_{\mathrm{y}i},
\end{equation}
with $\vek{w}_{\mathrm{y}i} \sim \mathcal{N}(\overline{\vek{y}}_{\mathrm{y}i}^{*}, (\vek{\Sigma}^*)_{\mathrm{y}i}^{2})$ being the weights of the output layer at time $t_i$. The components of $\vek{w}_{\mathrm{y}i}$ are computed in a similar manner as \refeq{eq:alphacompute} and \refeq{eq:variancecompute}. Note that input $ \boldsymbol{\mathit{\Psi}}_{i}(\vek{x}_i^*)$ contains the most probable value of $p(\vek{h}_{i})$.

\subsection{Uncertainty propagation}

With the propagation of the gate predictions $\hat{\vek{s}}_{\mathrm{g}i}$ through the LSTM cell (\refeq{lstm_transition}) and the output layer \refeq{predictionLSTM}), one obtains the predictive output $\hat{\vek{y}}_i$. To overcome the nonlinear activation functions (\refeq{eq:ARDLSTM}) we sample from the Gaussian ARD output $\vek{s}_{\mathrm{g}i}$, obtained from the given posterior weights $\vek{w}_{\mathrm{g}i}$ (\refeq{eq:ARDpredict}) by use of Monte Carlo simulation. After propagation through the gate activation function the mean and variance of the potentially non-Gaussian distribution can be obtained by:
\begin{equation}
\begin{aligned}
\vek{g}_{i}\vert_{\mu} &= \frac{1}{K} \sum_{K} (\sigma_\mathrm{g}(\mathcal{N}(\hat{\vek{s}}_{\mathrm{g}i})) \\
\vek{g}_{i}\vert_{\sigma} &= \frac{1}{K}\sum_{K} \mathrm{abs}(\sigma_\mathrm{g}(\mathcal{N}(\hat{\vek{s}}_{\mathrm{g}i})) - \vek{g}_{i}\vert_\mu).
\end{aligned}
\end{equation}
with hyperparameters $\vek{\alpha}_{\mathrm{g}i} \in \mathbb{R}^{(1+n_f+n_m) \times n_m}$ and $ \vek{\beta}_{\mathrm{g}i} \in \mathbb{R}^{M}$ the neural network output, the layerwise composition of the output prediction in \refeq{eq:outputPredict} can be formulated using \refeq{predictionLSTM}:
\begin{equation}\label{prediction2}
\hat{\vek{y}}_i(\vek{\alpha}_{\mathrm{y}i}, \vek{\beta}_{\mathrm{y}i},\vek{\alpha}_{\mathrm{g}i}, \vek{\beta}_{\mathrm{g}i})= \sigma_y(\vek{\alpha}_{\mathrm{y}i}, \vek{\beta}_{\mathrm{y}i})  \circ \sigma_{\hbar,\mathrm{MP}}^i(\vek{\alpha}_{\mathrm{g}i}, \vek{\beta}_{\mathrm{g}i}) \circ ... \circ \sigma_{\hbar,\mathrm{MP}}^1(\vek{\alpha}_{\mathrm{g}i}, \vek{\beta}_{\mathrm{g}i}).
\end{equation}

\section{Recurrent relevance determination}
\label{sec:relevance_determination_in_LSTM_cells}

To conform with the given data set $\vek{y}$ the hyperparameters for posterior predictive need to be iteratively re-estimated using \refeq{eq:alphacompute} and \refeq{eq:variancecompute}. This iterative procedure can be separated into initialization, forward propagation and backward propagation. An overview of the iterative procedure for the LSTM cell has been set out in Algorithm \ref{alg:ARDLSTMOpti}. The output layer is optimized similarly.
\begin{algorithm}[ht!]
\caption{ARD-LSTM optimization.}\label{alg:ARDLSTMOpti}
\begin{algorithmic}[1]
\Require{Input: $input \gets \in [-1, 1]$} \Comment{Normalized input}	
\Require{Prior for the weight: $\vek{\mu}_{\mathrm{g}i} \gets \mathcal{N}(0, \vek{\alpha}_{\mathrm{g}i}^{-1})$}
\Require{Likelihood hyperprior: $\vek{\beta}_{\mathrm{g}i} \gets \mathrm{Gamma}_{\mathrm{g}i}(\vek{\beta}_{\mathrm{g}i}\vert 0,0)$} \Comment{Uniform in $\mathrm{logarithmic}$ scale}
\Require{Hyperprior for the weight: $\vek{\alpha}_{\mathrm{g}i} \gets \mathrm{Gamma}_{\mathrm{g}i}(\vek{\alpha}_{\mathrm{g}i}\vert 0,0)$} 
\State Prior covariance $\vek{\Sigma}_{\mathrm{g}i} \gets diag(\vek{\alpha}_{\mathrm{g}i}^{-1})$
\While{$\mathcal{L}_{\mathrm{y}i}$ is not converged}  \Comment{Loops through epochs}	
\State $prediction(t) \gets input(t)$ \Comment{Forward propagation}
\State $\vek{y}(t) \gets prediction(t)$ \Comment{Backward propagation}
\State $\vek{\alpha}_{\mathrm{g}i} \gets \dfrac{\partial \mathcal{L}_{\mathrm{y}i}}{\partial \mathrm{log} \vek{\alpha}_{\mathrm{g}i}}$; \hspace{5mm} $\vek{\beta}_{\mathrm{g}i} \gets \dfrac{\partial \mathcal{L}_{\mathrm{y}i}}{\partial \mathrm{log} \vek{\beta}_{\mathrm{g}i}}$     
\State $\vek{\Sigma}_{\mathrm{g}i} \gets \mathcal{N}(\vek{w}_{\mathrm{g}i} \vert \vek{s}_{\mathrm{g}i},\vek{\alpha}_{\mathrm{g}i}, \vek{\beta}_{\mathrm{g}i})$           
\State $\vek{\mu}_{\mathrm{g}i} \gets \mathcal{N}(\vek{w}_{\mathrm{g}i} \vert \vek{s}_{\mathrm{g}i},\vek{\alpha}_{\mathrm{g}i}, \vek{\beta}_{\mathrm{g}i})\vert_{\mathrm{MP}} = \mathcal{N}(\vek{\mu}_{\mathrm{g}i}, \vek{\Sigma}_{\mathrm{g}i})\vert_{\mathrm{MP}}$
\EndWhile
\State \textbf{return} $\vek{\mu}_{\mathrm{g}i}, \vek{\Sigma}_{\mathrm{g}i}$\Comment{The posterior predictive}
\end{algorithmic}
\end{algorithm}

\subsection{Forward propagation}

To start the iterative procedure one has to initialize the prior over weights $\vek{w}_{\mathrm{g}i}$ which will serve as the basis for further optimization. Given the prior one can predict $\vek{s}_{\mathrm{g}i}$, and the network output by forward step: 

\begin{enumerate}
\item Define the Gaussian distributed zero-mean prior over $\vek{w}_{\mathrm{g}i}$ and  $\vek{w}_{\mathrm{y}i}$ using \refeq{eq:alpha_hyperprior}. Here, $\vek{\alpha}_{\mathrm{g}i}$ and $\vek{\alpha}_{\mathrm{y}i}$ are the (inverse) variance, described by and randomly sampled from a non-informative gamma distribution (in logarithmic scale). As many values tend to go to infinity, numerical instability must be taken care of. Therefore, values are bound to $\vek{\alpha}_{\mathrm{g}i} \in [1\num{e1}, 1\num{e6}]$. By setting a lower bound $>0$ to the $\alpha$ hyperparameter, a prior is imposed that motivates weight uniformity and therefore optimization stability (as $\alpha_{\mathrm{g}ikl} \rightarrow 0$ causes $\alpha_{\mathrm{g}ikl}^{-1}\rightarrow \infty$) throughout. It is important to note that for every individual weight, an individual hyperparameter is associated, which helps to exploit sparsity. For the first iterative step the mean is sampled from this prior only. As a result, the sampled mean will be rather small and the covariance will be dominated by $\vek{\alpha}$. This will promote sparsity at the onset of optimization.

\item Define the prior covariance $\vek{\Sigma}_{\mathrm{g}i}$ from $(\vek{\alpha}_{\mathrm{g}i})^{-1} \vek{I}$. Covariance $\vek{\Sigma}_{\mathrm{y}i}$ can be computed equally.
\end{enumerate}
One can now propagate the recurrent cell states over all time steps. 

\begin{enumerate}
\item\label{enum:forward_1} Initialize hidden state $\vek{h}_0 = 0$ and cell state $\vek{c}_0 = 0$ at $t_i=0$.
\item\label{enum:forward_2} Construct the gate input from $\boldsymbol{\mathit{\Phi}}_{i} = [\vek{x}_i,\vek{h}_i]$.
\item\label{enum:forward_3} Compute posterior predictive $\vek{s}_{\mathrm{g}i} = \boldsymbol{\mathit{\Phi}}_{i} \vek{\mu}_{\mathrm{g}i}$, defined for every gate individually. For the first iteration $\vek{\mu}_{\mathrm{g}i}$ is defined by prior samples, in subsequent steps by its posterior. This is the linear part of equation  \refeq{lstm_transition}. The corresponding predictive variance can be obtained by $(\vek{\sigma}_{\mathrm{g}i})^2 = (\boldsymbol{\mathit{\Phi}}_{i})^T \vek{\sigma}_{\mathrm{g}i} \boldsymbol{\mathit{\Phi}}_{i}$.
\item\label{enum:forward_4} Sample from Gaussian $\mathcal{N}\left(\vek{\mu}_{\mathrm{g}i}, \vek{\Sigma}_{\mathrm{g}i}^2 \right)$ and feed through the corresponding gate activation functions, giving the gate outputs $\vek{g}_{i}$ according to \refeq{lstm_transition}.
\item\label{enum:forward_5} Define $\vek{h}_{i,\mathrm{MP}}$ from the the most probable values of the sampled $\vek{h}_{i}$. This is both propagated to the output layer as propagated to the subsequent time step, alike traditional LSTM. Note that $\vek{C}_{i}$ is propagated in samples.
\end{enumerate}
After forward propagation through every time step, the hidden state $\vek{h}_{i,\mathrm{MP}}$ for every step is defined. The input for the output layer then yields $\boldsymbol{\mathit{\Psi}}_{\mathrm{y}i} = [ 1,\vek{h}_{i,\mathrm{MP}}]$. The posterior predictive of the output layer can now be computed from \refeq{eq:ARDpredict}.

\begin{figure}[ht!]
\centering
\includegraphics[width=0.47\textwidth]{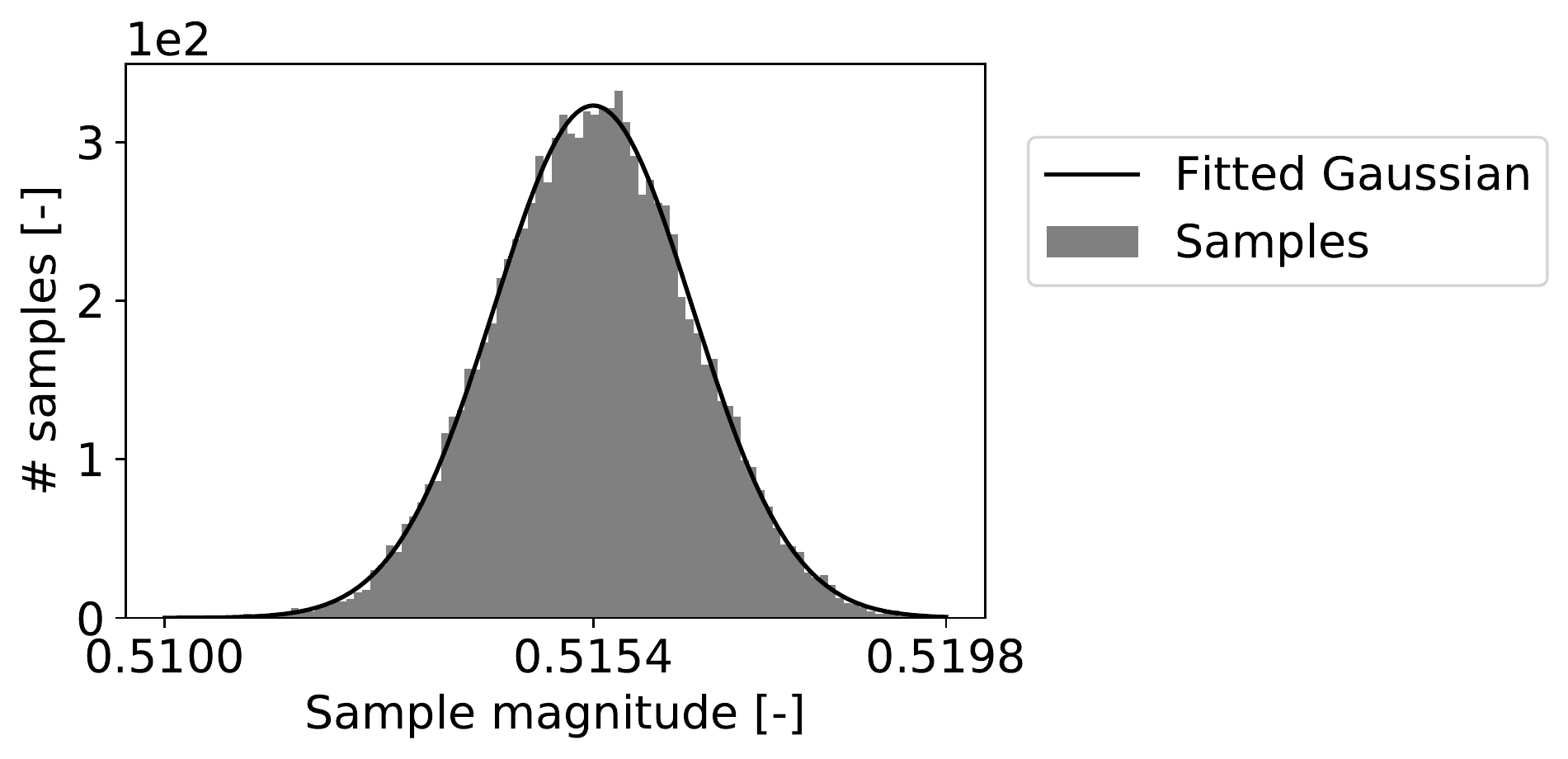}
\caption{Sampled state values and a fitted Gaussian probability density function of a single hidden state $\vek{h}_{ijl}$ unit $l$, selected by its largest magnitude over all entries.}
\label{fig:samples_fitted_gausian}
\end{figure}

An approximation of the unknown distribution over $\vek{h}_{i}$ is required to overcome expensive sampling procedures. Most-probable approximation by the mean value is reasonable for symmetric distributions representing the prior and posterior with a low data variance combined with a large semi-linear section in the hyperbolic tangent and sigmoid activation functions. This is confirmed by the posterior distribution over $\vek{h}_{i}$ given in \reffig{fig:samples_fitted_gausian}, obtained by taking \num{25000} samples of $\vek{h}_{i}$ after propagating the prior only. 

\subsection{Backward propagation}
Alike traditional LSTM, backward propagation is done by the assumption of small target data variance and thus by only using the mean-based gradient through the memory cell states. This yields an equal backward propagation as described by \cite{Gers.2000}. However, in the gates and the output layer one needs to incorporate the predictive uncertainty of the ARD framework to. With $\mathcal{L}_{\mathrm{g}i}$ denoting to the likelihood for every gate $\mathrm{g}$ (see \refeq{eq:marginal_likelihood_cost}), we denote to the output layer likelihood with $\mathcal{L}_{\mathrm{y}i}$. The continuous gradient from the likelihood to the hidden state is then given by $\dfrac{\partial \mathcal{L}_{\mathrm{y}i}}{\partial \vek{h}_{i, \mathrm{MP}}}$ and $\dfrac{\partial \mathcal{L}_{\mathrm{g}i}}{\partial \vek{h}_{i-1, \mathrm{MP}}}$. This is done by computing the partial derivatives of the likelihood objective function in  \refeq{eq:marginal_likelihood_cost}, giving:
\begin{equation}
\label{eq:ht_gradient_compute}
\begin{aligned}
\frac{\partial \mathcal{L}_{\mathrm{y}i}}{\partial \vek{h}_{i, \mathrm{MP}}} = \sum_{O} -\left[\vek{C}_{\mathrm{y}i} ^{-1} (\vek{\alpha}_{\mathrm{y}i}^{-1}\vek{I})\vek{h}_{i,\mathrm{MP}} - \vek{y}_i^T\left( \vek{C}_{\mathrm{y}i} ^{-2}(\vek{\alpha}_{\mathrm{y}i}\vek{I})\vek{h}_{i,\mathrm{MP}}\right) \vek{y}_i\right],
\end{aligned}
\end{equation}
with $\vek{C}_{\mathrm{y}i} = \vek{\beta}_{\mathrm{y}i}^{-1}\vek{I} + \vek{h}_{i,\mathrm{MP}} (\vek{\alpha}_{\mathrm{y}i}^{-1}\vek{I})\vek{h}_{i,\mathrm{MP}}^T$ and $O$ being the width of the output layer. This yields:
\begin{equation}
\frac{\partial \mathcal{L}_{\mathrm{y}i}}{\partial \vek{h}_{i, \mathrm{MP}}} = \sum_{O} \left( \vek{C}_{\mathrm{y}i} ^{-1} \vek{y}_i \vek{y}_i^T  \vek{C}_{\mathrm{y}i} ^{-1} -  \vek{C}_{\mathrm{y}i} ^{-1} \right) \vek{h}_{i, \mathrm{MP}} \vek{\alpha}_{\mathrm{y}i}^{-1}\vek{I}.
\end{equation}
Similarly for all LSTM gates:
\begin{equation}
\frac{\partial \mathcal{L}_{\mathrm{g}i}}{\partial \vek{h}_{i-1, \mathrm{MP}}} = \sum_M \left( \vek{C}_{\mathrm{g}i}^{-1}  \vek{s}_{\mathrm{g}i} \vek{s}_{\mathrm{g}i}^T  \vek{C}_{\mathrm{g}i}^{-1} -  \vek{C}_{\mathrm{g}i}^{-1} \right) \vek{h}_{i-1, \mathrm{MP}} \vek{\alpha}_{\mathrm{g}i}^{-1}\vek{I},
\end{equation}
with $\vek{C}_{\mathrm{g}i}  = \beta_i^{-1}\vek{I} + \vek{h}_{i-1, \mathrm{MP}} ( \vek{\alpha}_{\mathrm{g}i}^{-1}\vek{I})\vek{h}_{i-1, \mathrm{MP}}^T$.

Accordingly, the backward propagation steps can be described:
\begin{enumerate}
\item\label{enum:back_1} Obtain $\dfrac{\partial \mathcal{L}_{\mathrm{y}i}}{\partial \vek{h}i}$ using  \refeq{eq:ht_gradient_compute}.
\item\label{enum:back_2} Compute the LSTM mean-based gradients according to  \refeq{eq:LSTM_backprop}, with $\delta \vek{h}_i=0$ and \\ $\partial \vek{C}_{\mathrm{y},i+1}=0$ at $i=m$. Note that although the gradient computation the cell states is mean-based, the gate outputs are represented by a complicated distribution function, originated from the predictive mean and variance of a Gaussian distributed output (forward propagation step \ref{enum:forward_3} and \ref{enum:forward_4}).
\item\label{enum:back_3} Compute the mean of the propagated gradient samples and use the mean-based predictions as performed in Forward propagation step \ref{enum:forward_3} to update the target function  $\vek{s}_{\mathrm{g}i}$ according to  \refeq{eq:target_update}. Only the mean target is relevant here, as predictions are performed around this very mean. A non-restricted gradient update scheme could lead to instabilities. An ADAM optimization scheme can be applied to stabilize  \refeq{eq:target_update}. To prevent further numerical instabilities with gradient inverses, the target functions are bound between $[\num{-9}, \num{9}]$ for sigmoid functions and $[\num{-5}, \num{5}]$ for hyperbolic tangent activation functions. 
\item\label{enum:back_4} Compute the gradient for each individual gate likelihood objective to the hidden state $\dfrac{\partial\vek{s}_{\mathrm{g}i}}{\partial \vek{h}_{i-1, \mathrm{MP}}}$ and repeat backward propagation steps \ref{enum:back_1}-\ref{enum:back_5} up to $i=0$. 
\item\label{enum:back_5} Update $\vek{\alpha}_{\mathrm{g}i}$, $\vek{\alpha}_{\mathrm{y}i}$, $\vek{\beta}_{\mathrm{g}i}$ and $\vek{\beta}_{\mathrm{y}i}$ using  \refeq{eq:alphacompute} and \refeq{eq:variancecompute}.
\item\label{enum:back_6} Compute the predictive mean $\vek{\mu}_{\mathrm{g}i}$, $\vek{\mu}_{\mathrm{y}i}$ and covariance $\vek{\Sigma}_{\mathrm{g}i}$ and $\vek{\Sigma}_{\mathrm{y}i}$ using  \refeq{eq:musigma}.
\item\label{enum:back_7}  Compute $\vek{\gamma}_{\mathrm{g}i}$ and $\vek{\gamma}_{\mathrm{y}i}$ according to \refeq{eq:pruning_setup}. This can be used to update the hyperparameters $\vek{\alpha}_{\mathrm{g}i}$, $\vek{\alpha}_{\mathrm{y}i}$, $\vek{\beta}_{\mathrm{g}i}$ and $\vek{\beta}_{\mathrm{y}i}$ at the next iteration step and prune  $\vek{\mu}_{\mathrm{g}i}$ and $\vek{\mu}_{\mathrm{y}i}$ at the current step.

\end{enumerate}
Repeat the forward- and backward propagation steps until convergence, when a maximum of the marginal likelihood is attained. To ensure proper convergence a criterion over multiple iteration steps is defined. Convergence is assumed and the optimization is terminated at iteration step $n$ when the criterion
\begin{equation}
\vert \mathcal{L}_{\mathrm{y}, n-20} - \mathcal{L}_{\mathrm{y},n} \vert \leq \num{2e-2}
\end{equation}
has been met twice.

Note that $\vek{s}_{\mathrm{g}i}$ needs to be iteratively updated via:
\begin{equation}
\label{eq:target_update}
\vek{s}_{\mathrm{g}i} = \frac{1}{K}\sum_{K} \hat{\vek{s}}_{\mathrm{g}i} + \lambda \frac{\partial \boldsymbol{\mathcal{L}_{\mathrm{g}i} }}{\partial \hat{\vek{s}}_{\mathrm{g}i}},
\end{equation}
in which $\dfrac{\partial \boldsymbol{\mathcal{L}_{\mathrm{g}i}}}{\partial \hat{\vek{s}}_{\mathrm{g}i}}$ is the corresponding Jacobian, $\hat{\vek{s}}_{\mathrm{g}i}$ is the posterior predictive, and $K$ is the number of Monte-Carlo samples taken of $\hat{\vek{s}}_{\mathrm{g}i}$. As the gradient results from both the prediction at the current time step and the recurrent relation, the likelihood must be computed not only over the prediction at $t_i$, but also on the gates at time $t_{i+1}$.

\subsection{Pruning threshold}
\label{pruning_threshold}
With the assumptions made on the hyperparameter boundaries and the resulting necessity for a less strict pruning requirement, a threshold parameter $\tau$ is introduced. Adequate predictive results have been found for a relatively conservative $\tau = \num{1e-4}$. Pruning is performed where the covariance $\vek{\Sigma}$ is primarily constrained by $\vek{\alpha}_{\mathrm{g}i}$, thus where $\gamma_{\mathrm{g}ikl} \leq \tau$. By increasing $\tau$ one may obtain a higher sparsity percentage as the weight is required to be less constrained by the prior. This will naturally affect its predictability.

Input $[ \vek{x}_i, \vek{h}_i ]$ for each ARD-LSTM gate is normalized and restricted by a hyperbolic tangent, hence $\boldsymbol{\mathit{\Phi}} \in [-1,1]$. Additionally, bounds of the hyperparameters are $\vek{\alpha}_{\mathrm{g}i} \in [\num{1e1}, \num{1e6}]$ and $\vek{\beta}_{\mathrm{g}i} \in [\num{1e4}, \num{1e6}]$.  In practice, many values in $\vek{A}$ will tend to go to infinity and will hence be bounded to $\num{1e6}$. To strive dominance of $\vek{\alpha}_{\mathrm{g}i}$, $\vek{\Sigma}_{\mathrm{g}i}$ must be primarily described by its diagonal. This dominance is promoted by the high bounds on $\vek{\beta}_{\mathrm{g}i}$ and insignificant recurrent input throughput, allowing for insignificant covariances away from the diagonal.

An alternative approach to pruning would be to evaluate $\vek{\alpha}_{\mathrm{g}i}$ only and prune for any of upper boundary values in $\vek{\alpha}_{\mathrm{g}i}$ (if $\alpha_{\mathrm{g}ikl} \rightarrow \infty$ then $\mu_{\mathrm{g}ikl} \rightarrow 0$) \cite{Shutin.2011, Tipping.2001}. This bi-directional update scheme on both the linearized target $\vek{s}_{\mathrm{g}}$ and hyperparameters has been found to be unstable for unbounded hyperparameters, causing a numerically singular Hessian matrix $\vek{H}_{\mathrm{g}i} = \vek{\beta}_{\mathrm{g}i} \odot \boldsymbol{\mathit{\Phi}}_i^T \boldsymbol{\mathit{\Phi}}_i + \vek{\alpha}_{\mathrm{g}i} \vek{I}$, as also hinted by \cite{Tipping.2001}. In regular relevance vector machine the input and output sizes are equal. To prevent ill-conditioning in regular relevance vector machine one may remove corresponding basis functions (inputs) from $\boldsymbol{\mathit{\Phi}}_i$.

In ARD-LSTM, with $\vek{w}_{\mathrm{g}i} \in \mathbb{R}^{(1+n_f+n_m) \times n_m}$ the input dimension $(1+n_f+n_m)$ is broadcast over all $n_m$ network units  and over all $N$ basis functions in the subsequent output layer. One may be able to prune corresponding inputs if and only if the input $\boldsymbol{\mathit{\Phi}}_i$ is redundant for all broadcast widths. Due to this unfeasible requirement, the input pruning will be very insignificant and therefore not preventing any ill-conditioning. With an upper limit to $\vek{A}$ ill-conditioning can also be prevented, but pruning for $\alpha_{\mathrm{g}ikl} \rightarrow \infty$ is disregarded. 

A similar approach on the threshold, as performed for the gates $\mathrm{g}$, is also followed for the output layer $\mathrm{y}$.

\section{Structural application}
\subsection{Constitutive model description}

\begin{figure}[ht!]
\centering
\includegraphics[width=0.7\textwidth]{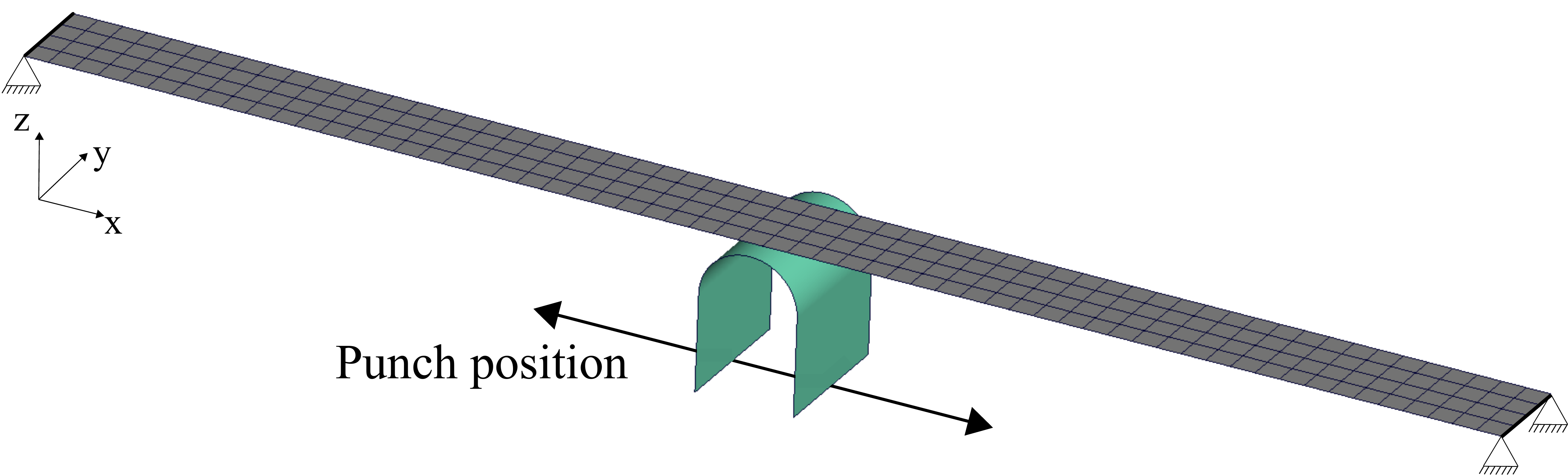}
\caption{Bending test overview, with indicated the varying punch position along the longitudinal axis.}
\label{fig:BE_doe}
\end{figure}

\begin{figure}[ht!]
\centering
\includegraphics[width=0.8\textwidth]{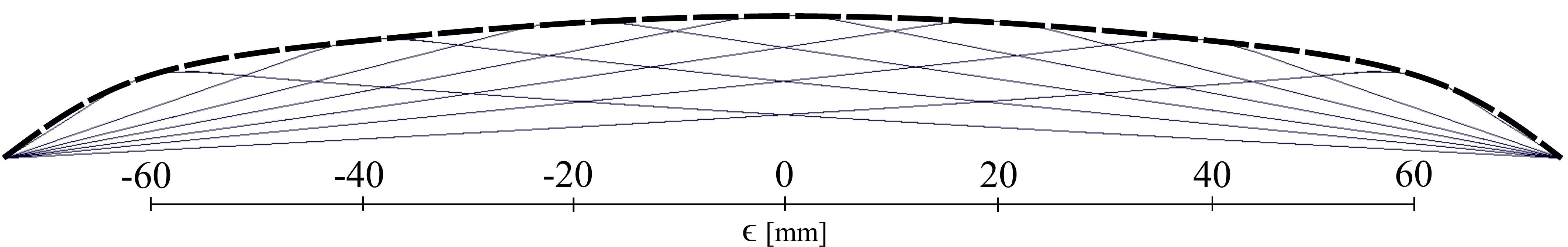}
\caption{Bending test DOE location indication and overlaying convex envelope.}
\label{fig:BE_envelope}
\end{figure}

The data set comes from a bending test using a relatively simple \SI{1.5}{\milli\meter} thick and \SI{150}{\milli\meter} long metal specimen strip, see \reffig{fig:BE_doe}. The strip is fixed using rotational bearings on both lateral sides. The punch is given a constant mass of \SI{10}{\kilogram} and an initial impact speed of \SI{2.0}{\meter\per\second}. Additionally, the simulation time is set to \SI{20}{\milli\second} with an output frequency of \SI{2}{\hertz}, hence $T=\num{41}$ time steps. The goal is to predict the nodal displacement ($x,y,z$) using the data present on a total of \num{7} samples -~in FEM better known as designs of experiments (DOEs)~-, obtained from a uniform distribution of the punch position over the given parameter space. 
In \reftab{tab:DOEspec} the DOEs are chosen within the parameter space $\epsilon \in [\SI{-60}{\milli\meter},\SI{60}{\milli\meter}] $, with the longitudinal mid as specimen center. Due to the punch impact lower frequency oscillations are found in the deformed specimen after reaching maximum deformation. In \reffig{fig:BE_envelope} the deformation of all DOEs is visualized in $xz$-plane, indicating a convex envelope that describes maximum deformations for arbitrary punch positions.
\begin{table}[ht!]
\centering
\caption{DOE specification.}
\label{tab:DOEspec}
\begin{tabular}{@{}r|ccccccc@{}}
\toprule
DOE number          & \num{1}   & \num{2}   & \num{3}   & \num{4} & \num{5}  & \num{6}  & \num{7}  \\ \midrule
$\epsilon \quad [\SI{}{\milli\meter}]$& \num{-60} & \num{-40} & \num{-30} & \num{0} & \num{20} & \num{40} & \num{60} \\ \bottomrule
\end{tabular}
\end{table}
The punch is considered rigid and is limited to a displacement in $z$-direction only. The material used in the specimen is a 22MnB5 hot-forming steel. The material is modeled by an isotropic elasto-plastic strain hardening according to a Hershey yield criterion. A more elaborated description of the applied fundamental material model is given in \cite{Eller.2016}. The described model is implemented and solved using the explicit crash solver Virtual Performance Solution (VPS) \cite{Group.2020}. 

The constant mesh topology contains \num{305} nodes, and hence a total of $O=\num{915}$ ($x,y,z$) displacement output values need to be predicted per time step, giving $\vek{y} \in \mathbb{R}^{7 \times 41\times 915}$.

A variety of ARD-LSTM widths will be analyzed ($n_m \in \left\lbrace 16, 32, 64, 128\right\rbrace$) to compare the sparsity effect and convergence with traditional point estimate LSTM in varying network complexities. For ARD-LSTM the ADAM optimizer learning rate is set to $\lambda=0.005$ and a maximum number of \num{4000} epochs is considered. To allow for proper comparison, initialization of both traditional- and ARD-LSTM is performed using the same sampling on the hyperpriors and thus ensures an equal initial weight (mean). Additionally, as similar noise is expected over all data, hyperparameter $\vek{\beta}_{\mathrm{g}i}$ is initialized in the domain $\vek{\beta}_{\mathrm{g}i} \in [1\num{e4}, 1\num{e5}]$. During optimization the hyperparameter boundaries are set by $\vek{\alpha}_{\mathrm{g}i}  \in [1\num{e1}, 1\num{e6}]$ and $\vek{\beta}_{\mathrm{g}i}  \in [1\num{e4}, 1\num{e6}]$, $K=\num{100}$ samples are taken of $\hat{\vek{s}}_{\mathrm{g}i}$ and pruning is applied for $\gamma_k \leq 1\num{e-4}$.
This less strict pruning threshold has been set to conform to the imposed boundaries on $\vek{\alpha}_{\mathrm{g}i}$ and $\vek{\beta}_{\mathrm{g}i}$, which disallows $\alpha_{ikl} \rightarrow \infty$ and therefore demotivates $\Sigma_{ikkl} \approx \alpha_{ikl}^{-1}$ ( \refeq{eq:pruning_setup}, Section \ref{pruning_threshold}). Note that this threshold still ensures that the posterior covariance can be strongly dominated by the prior. The point estimate network has no weight regularization and is optimized over a single batch. Both network types have been created and optimized using the Tensorflow $v2.1.0$ framework \cite{Tensorflow.2015} on a single Nvidia Quadro P5000 GPU.

\subsection{Predictability assessment}

\subsubsection{Optimization}
\begin{figure}[ht!]
\centering
\includegraphics[width=0.5\textwidth]{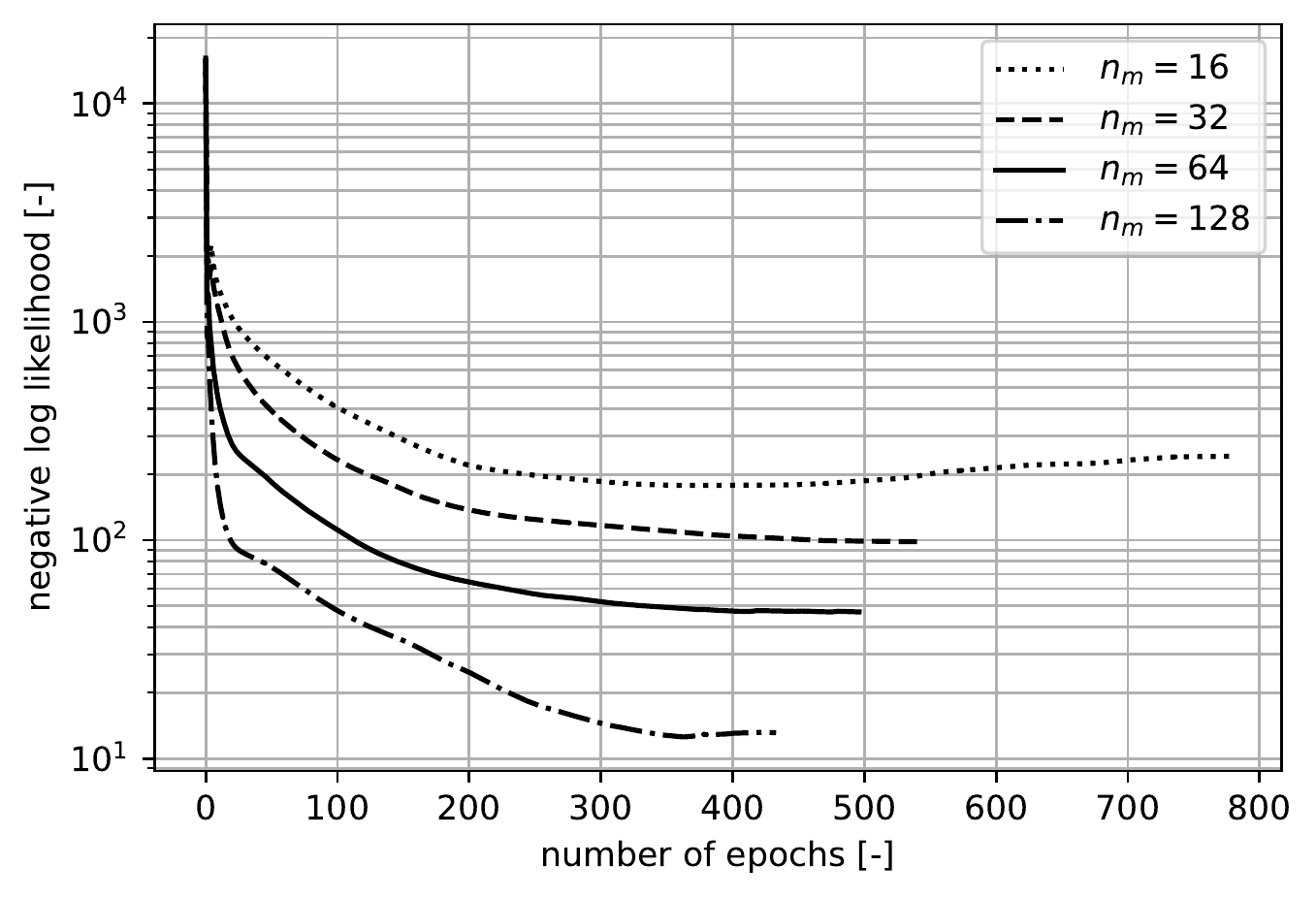}
\caption{Negative log likelihood optimization up to convergence.}
\label{fig:loss_curves}
\end{figure}

In \reffig{fig:loss_curves} the negative log likelihood for all given widths $n_m$ in ARD-LSTM is plotted. Whereas the larger networks are converging steadily, the smaller one for $n_m=16$ one obtains a noisy estimate. This indicates unterfitting on the given data set.
An overview of the optimization metrics is given in \reftab{tab:opti_metrics}. Due to the increase in the number of free variables variables ARD-LSTM requires a  longer computational time overall as well as per epoch. Although ARD-LSTM may be more memory intensive, it converges faster and hence does not require as many epochs as the classical LSTM. Comparison of ARD-LSTM and the point estimate LSTM is done over the coefficient of determination $\mathrm{R}^2$ ($\in [0,1]$):
\begin{equation}
\mathrm{R}^2 = 1 - \frac{\sum_i^m \sum_j^{n_b} \left(\vek{y}_{ij} - \frac{1}{{n_b}}\sum_{n_b} ({\vek{y}_{i}})^* \right)^2}{ \sum_i^m \sum_j^{n_b} \left(  \vek{y}_{ij}-\frac{1}{{n_b}}\sum_{n_b} {\vek{y}_{i}} \right)^2},
\end{equation}
where $\vek{y}_{ij}^*$ is approximated by the predictive mean (see \refeq{eq:ARDpredict}). For ARD-LSTM higher $\mathrm{R}^2$ values over all widths are found, correlating to a lower squared error. Whereas ARD-LSTM is able to cover $\SI{99.5}{\percent}$ of data variance at $n_m=32$, the point estimate framework only reaches similar result at $n_m=128$. For ARD-LSTM a significant increase in time per epoch is found for $n_m=128$. the cause of this is the required batch-wise computation of the covariance to fit the available memory.
\begin{table}[ht!]
\centering
\caption{Optimization metrics for all considered network widths ($n_m \in \left\lbrace 16, 32, 64, 128\right\rbrace$).}
\label{tab:opti_metrics}
\begin{adjustbox}{max width=\textwidth}
\begin{tabular}{@{}rccccc@{}}
\toprule
  & \multicolumn{1}{c}{$n_m$} & \multicolumn{1}{l}{epochs $[-]$} & \multicolumn{1}{l}{time $[\SI{}{\second}]$} & \multicolumn{1}{l}{time per epoch $[\SI{}{\second}]$}  & $\mathrm{R}^2$ $[-]$\\ \midrule
\multicolumn{1}{c}{\multirow{4}{*}{Point estimate LSTM}} & \multicolumn{1}{c|}{16}       & \num{4000}& \num{189}  & \num{0.047} & \num{0.994}\\
\multicolumn{1}{c}{}                                     & \multicolumn{1}{c|}{32}       & \num{4000}& \num{189}  & \num{0.048}  & \num{0.993}\\
\multicolumn{1}{c}{}                                     & \multicolumn{1}{c|}{64}       & \num{4000}& \num{190}  & \num{0.048}  & \num{0.994}\\
\multicolumn{1}{c}{}                                     & \multicolumn{1}{c|}{128}      & \num{4000}& \num{191}  & \num{0.048} & \num{0.996}\\ \midrule
\multirow{4}{*}{ARD-LSTM}                                & \multicolumn{1}{c|}{16}       & \num{779}& \num{153}  & \num{0.20} & \num{0.993}\\
                                                         & \multicolumn{1}{c|}{32}       & \num{541}& \num{151}  & \num{0.28} & \num{0.995}\\
                                                         & \multicolumn{1}{c|}{64}       & \num{497}& \num{317}  & \num{0.64}   & \num{0.998}  \\
                                                         & \multicolumn{1}{c|}{128}      & \num{434}& \num{1403}  & \num{3.23}  & \num{0.998}\\ \bottomrule
\end{tabular}
\end{adjustbox}
\end{table}

\begin{figure}[ht!]
\centering
\begin{subfigure}[c]{0.85\textwidth}
\centering
\includegraphics[width=\textwidth]{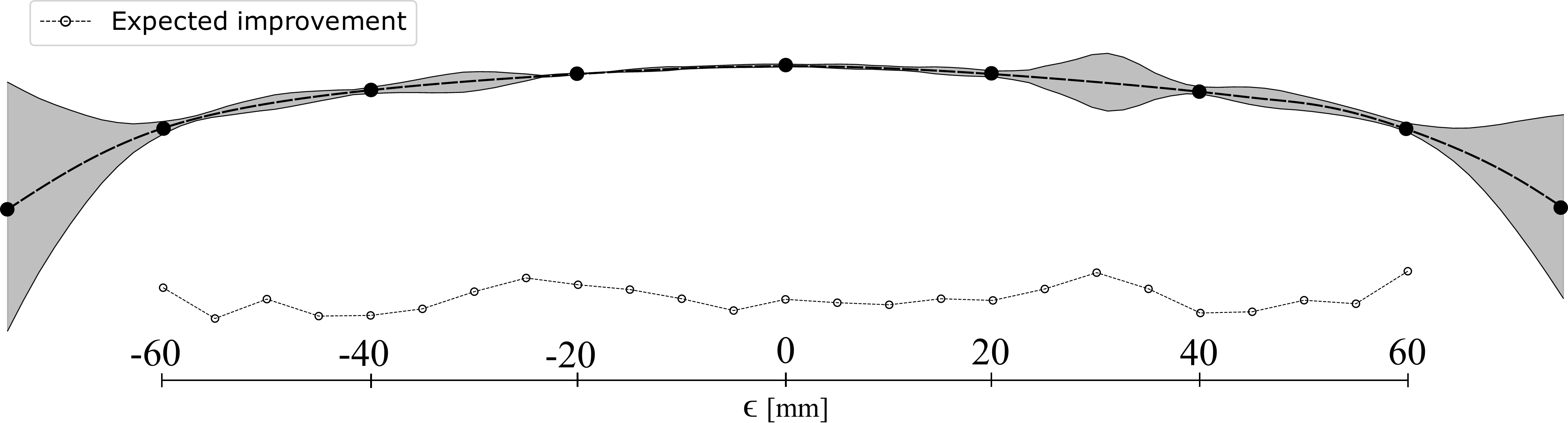}
\subcaption{Normalized standard deviation overlay of ARD-LSTM with $n_m=\num{32}$ trained on all DOEs.}
\label{fig:BE_envelope_overlay_base}
\end{subfigure}
\begin{subfigure}[c]{0.85\textwidth}
\centering
\includegraphics[width=\textwidth]{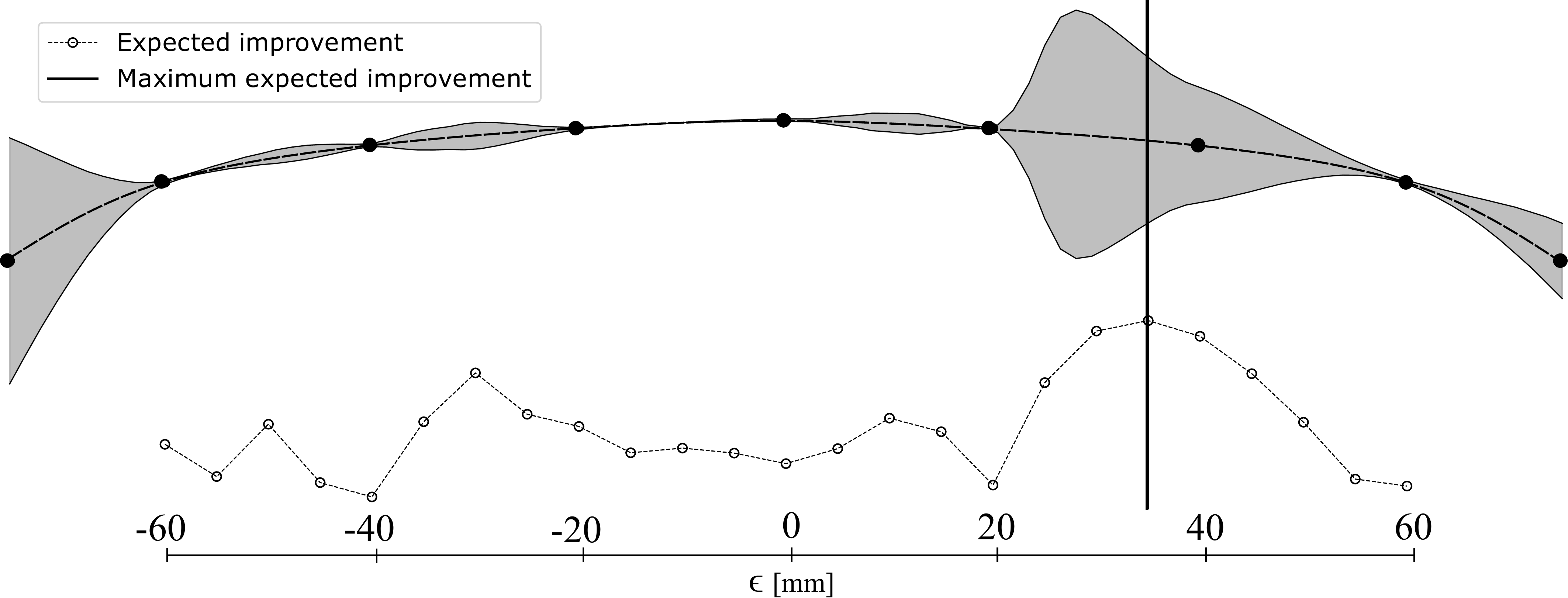}
\subcaption{Normalized standard deviation overlay of ARD-LSTM with $n_m=\num{32}$ trained without DOE \num{6} ($\epsilon=\SI{40}{\milli\meter}$).}
\label{fig:BE_envelope_overlay_vali}
\end{subfigure}
\caption{Bending test DOE location indication and overlaying normalized maximum standard deviation plots on the convex envelope described in \reffig{fig:BE_envelope}.}
\label{fig:BE_envelope_overlays}
\end{figure}

In \reffig{fig:BE_envelope_overlay_base} the normalized maximum predictive standard deviation $\sigma$ for $n_m=32$ over all time steps is plotted as an overlay on the convex envelope of \reffig{fig:BE_envelope}. The input $\epsilon$ is sampled \num{100} times linearly over the range $\epsilon \in [\SI{-75}{\milli\meter},\SI{75}{\milli\meter}]$. Hence, all values outside of this interval are extrapolated. Although the magnitude of the normalized predictive standard deviation is small for $n_m=32$ (see \reffig{fig:BE_envelope_overlay_base}), ARD-LSTM still clearly captures the difference in uncertainty between training samples, interpolation and extrapolation. A secondary overlay is given in \reffig{fig:BE_envelope_overlay_vali}, here of a network with $n_m=32$ that has been trained with without DOE \num{6} ($\epsilon = \SI{40}{\milli\meter}$, \reftab{tab:DOEspec}) in the training set. Due to the lack of domain knowledge around $\epsilon = \SI{40}{\milli\meter}$ the predictive uncertainty strongly increases and a local standard deviation increase from $\sigma = \num{1.8e-2}$ to $\sigma = \num{5.2e-2}$ is found. This corresponds to a predictive standard deviation of $\SI{0.88}{\milli\meter}$, where a predictive mean displacement of $\SI{7.03}{\milli\meter}$ is found.

With an eye on expensive training data generation, one desires to minimize the number of DOEs required. It may therefore be useful to analyze the location where the maximum expected improvement is expected. At every $\epsilon$ the expected improvement acquisition function can be expressed as \cite{Jones.2001}:
\begin{equation}
\mathrm{EI}(\epsilon) = \mathbb{E}\left[\mathrm{max}\left((y(\epsilon)-\hat{y}^*(\epsilon),0 \right)\right]
\end{equation}
in which $y(\epsilon)$ is given by \refeq{final_prediction}. Target $y \sim \mathcal{N}(\overline{y}, \sigma^2)$ and $\hat{y}^* \sim \mathcal{N}(\overline{y}^*, (\sigma^*)^2)$ for respectively the DOE output $y$ and predictive output $\hat{y}^*$, see \refeq{eq:pred_overview}. We desire to find DOE $\epsilon$ where a maximum expected improvement can be found. This yields for $\sigma^*(\epsilon) > 0$:
\begin{equation}
\mathrm{EI}(\epsilon) = \left(\overline{y}^*(\epsilon) - y\right)\varpi \left(\frac{\overline{y}^*(\epsilon) - y}{\sigma^*(\epsilon)}\right) + \sigma^*(\epsilon)\mathcal{N}\left(\frac{\overline{y}^*(\epsilon) - y}{\sigma^*(\epsilon)}\right)
\end{equation}
with $\varpi$ the cumulative distribution function of the standard Gaussian distribution. For more insight $\num{25}$ linearly spread designs are considered over the given domain $\epsilon \in \left[ \SI{-60}{\milli\meter}, \SI{60}{\milli\meter}\right]$ and added in \reffig{fig:BE_envelope_overlays}, below the variance overlay on the convex envelope. As a result of an increased predictive uncertainty (see \reffig{fig:BE_envelope_overlay_vali}), the maximum expected improvement is found for $\epsilon=\SI{35}{\milli\meter}$, which is close to the removed DOE at $\epsilon=\SI{40}{\milli\meter}$. Hence, application of ARD-LSTM analysis provides the additional benefit of obtaining domain knowledge on the basis of predictive uncertainty. This domain knowledge allows the selection of a minimal-sized data set for satisfying interpolation uncertainty and thus model generalization.

\subsubsection{Perspectives on sparsity}
\begin{figure}[ht!]
\centering
\begin{subfigure}[c]{0.47\textwidth}
\centering
\includegraphics[width=\textwidth]{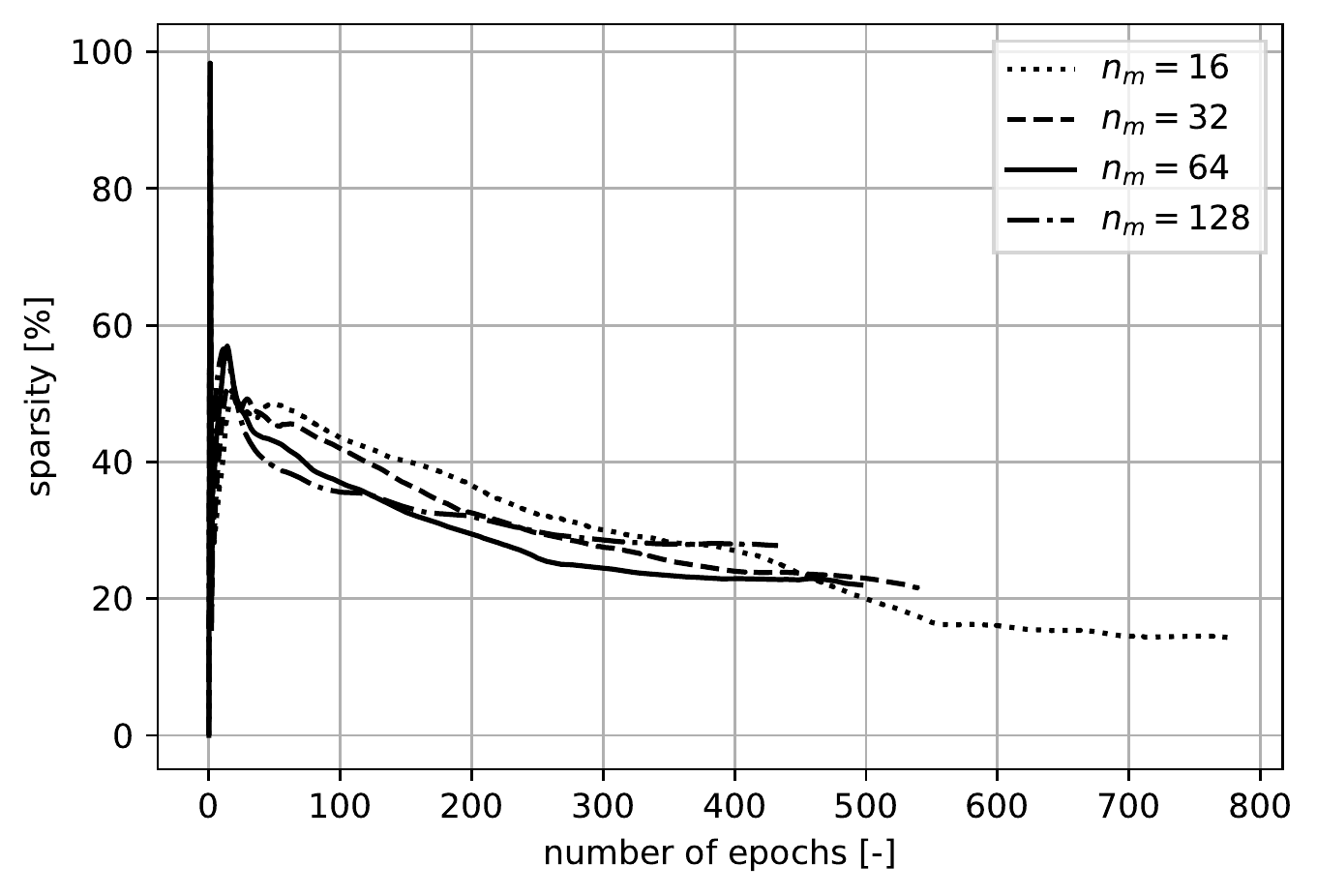}
\subcaption{Dense layer sparsity.}
\label{fig:Sparsity_Dense}
\end{subfigure}
\begin{subfigure}[c]{0.47\textwidth}
\centering
\includegraphics[width=\textwidth]{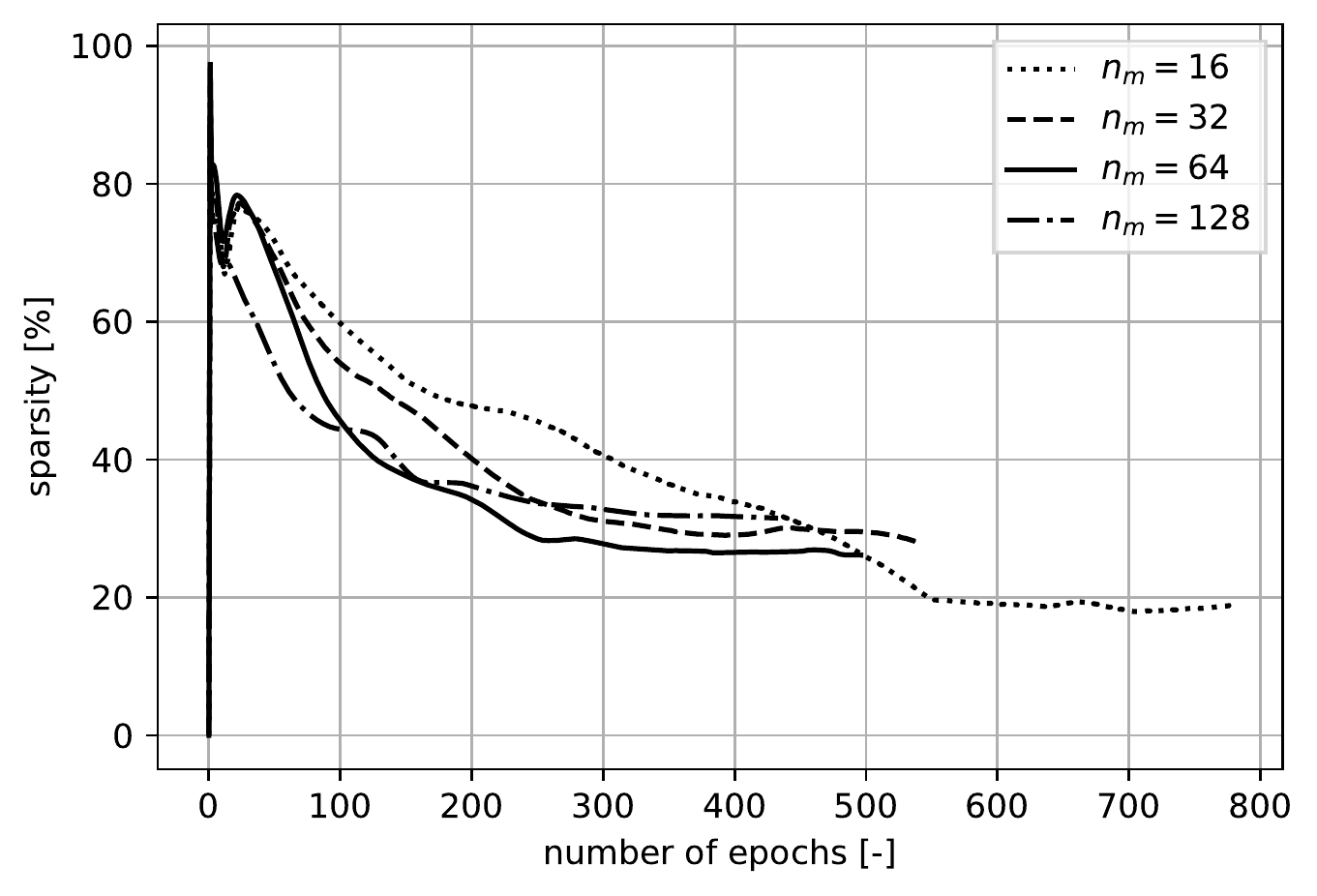}
\subcaption{LSTM cell sparsity.}
\label{fig:Sparsity_gates}
\end{subfigure}
\caption{Sparsity percentage for the ARD-LSTM cell and output layer.}
\label{fig:Sparsity}
\end{figure}

Sparsity is heavily stimulated at the onset of ARD-LSTM hyperparameter optimization, caused by the prior-driven $\gamma$ in initialization, see \refeq{eq:pruning_setup} and Section \ref{sec:relevance_determination_in_LSTM_cells}. In \reffig{fig:Sparsity} the convergence of weights sparsity in the output layer and LSTM cell is shown for different architectures denoted by $n_m$. One can see that starting from a high sparsity percentage, the ARD framework re-introduces weights according to their relevance.

A decreasing sparsity percentage for a more complex architecture may seem counter-intuitive. However, with a varying network complexity a different optimization objective is obtained. Note that sparsity in smaller networks may still be present due to weight redundancy.

\begin{figure}[ht!]
\centering
\includegraphics[width=0.9\textwidth]{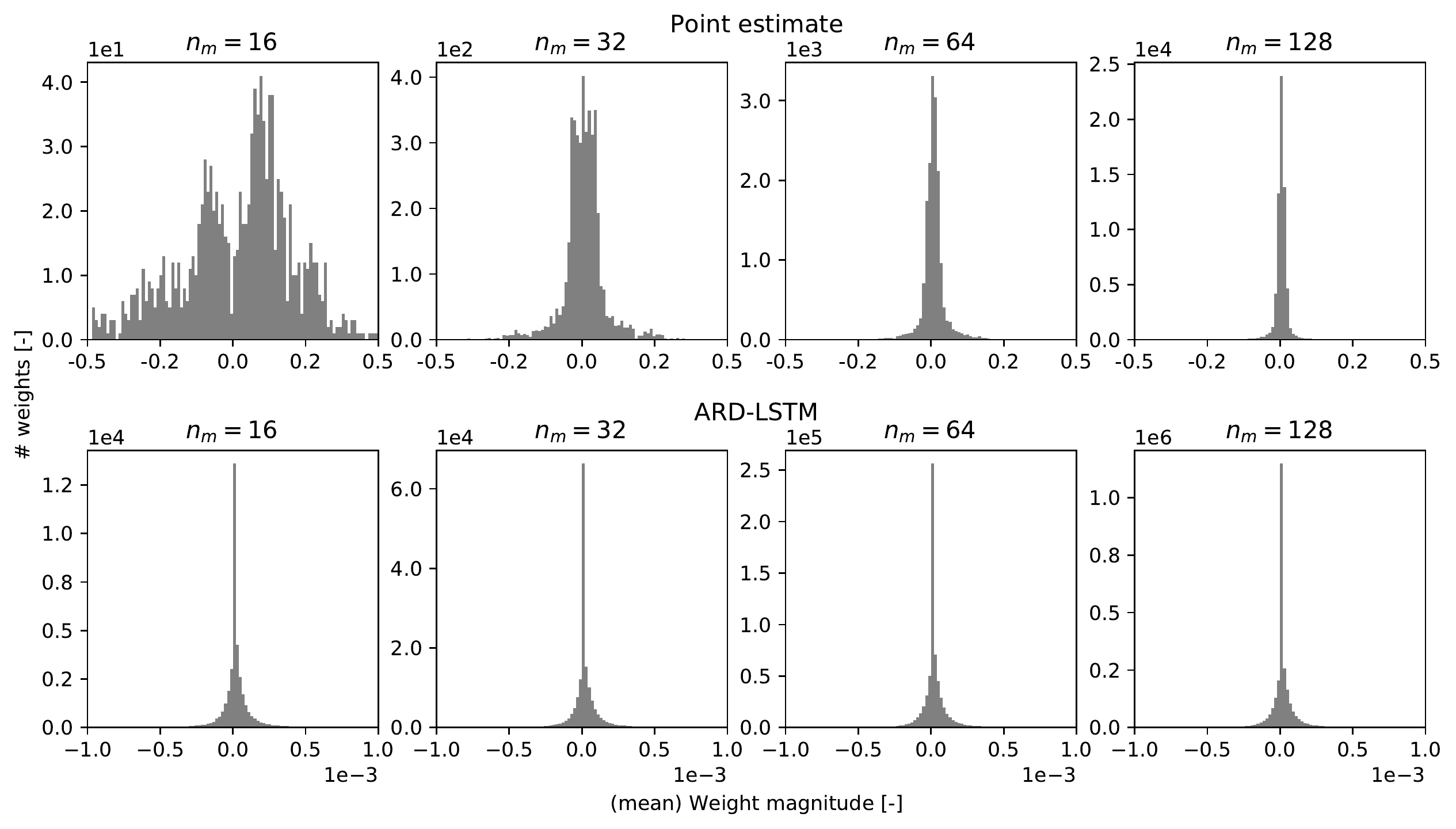}
\caption{LSTM cell weight magnitude for the point estimate LSTM (top row) and ARD-LSTM (bottom row).}
\label{fig:abs_weight}
\end{figure}

The (mean) weight magnitudes of ARD-LSTM and the corresponding point-estimate weights are depicted in \reffig{fig:abs_weight}. The set of weights in ARD-LSTM are all Laplace-like shaped. The point estimate LSTM only gets small-valued weights on larger network widths. Although the point estimate weights are valued around zero, none exactly equal it.  

\subsubsection{Perspectives on sampling}
The computational expenses for ARD-LSTM are significant. The covariance computation is in the order $\Theta(m \times N \times n_m^3)$ and storage in the order $\Theta(m \times N \times n_m^2)$, with $N$ being the width of the output layer or the LSTM gate output (for which $N = n_m$). Increasing the network width $n_m$ therefore leads to a significant increase in computational effort, also found in the optimization times in \reftab{tab:opti_metrics}. Furthermore, sampling of the posterior weights in the LSTM cell increases the storage by order $\Theta(S)$. With many samples required (typically $S>10^2$) to properly describe a complex distribution this may increase storage expenses significantly. The hidden state can, however, be approximated by a Gaussian (Section \ref{sec:relevance_determination_in_LSTM_cells}) and the cell state was found to be described similarly as well. One therefore may argue that only $S<10^2$ samples are required to properly describe these states. Still, sampling would increase storage significantly.

\begin{figure}[ht!]
\centering
\begin{subfigure}[c]{0.24\textwidth}
\centering
\includegraphics[width=\textwidth]{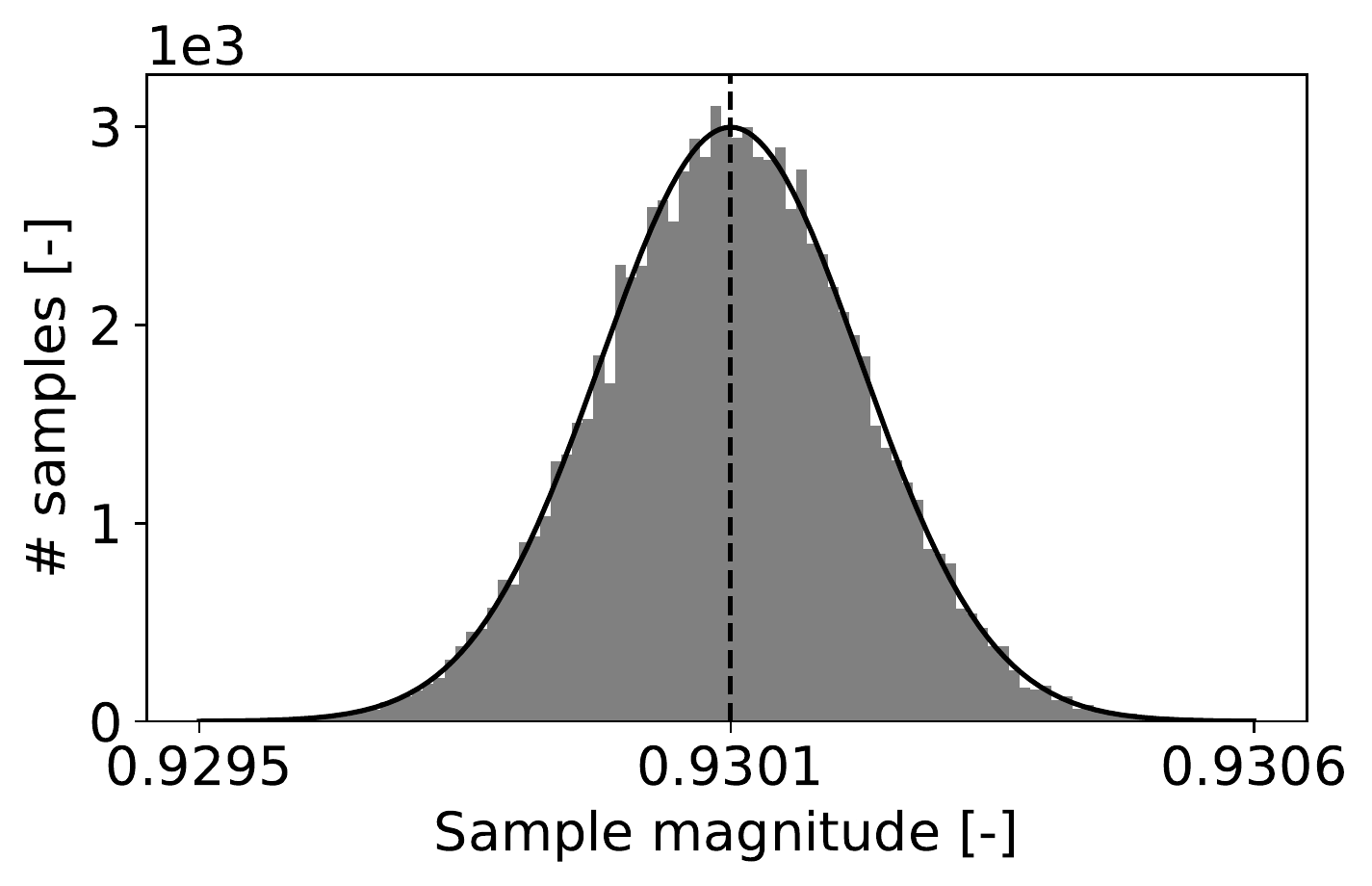}
\subcaption{$n_m=16$}
\label{fig:sampling_mean_posterior_16}
\end{subfigure}
\begin{subfigure}[c]{0.24\textwidth}
\centering
\includegraphics[width=\textwidth]{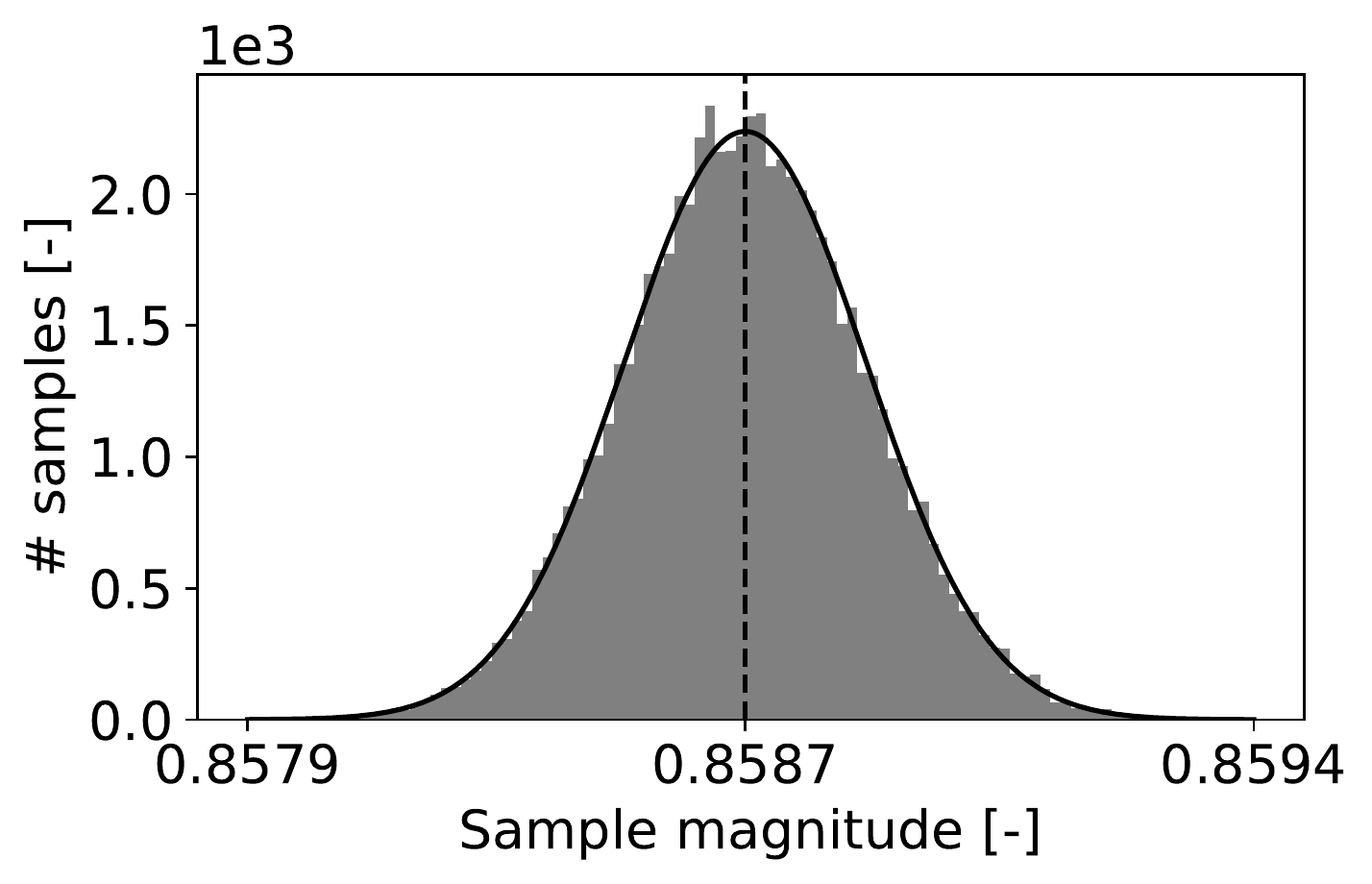}
\subcaption{$n_m=32$}
\label{fig:sampling_mean_posterior_32}
\end{subfigure}
\begin{subfigure}[c]{0.24\textwidth}
\centering
\includegraphics[width=\textwidth]{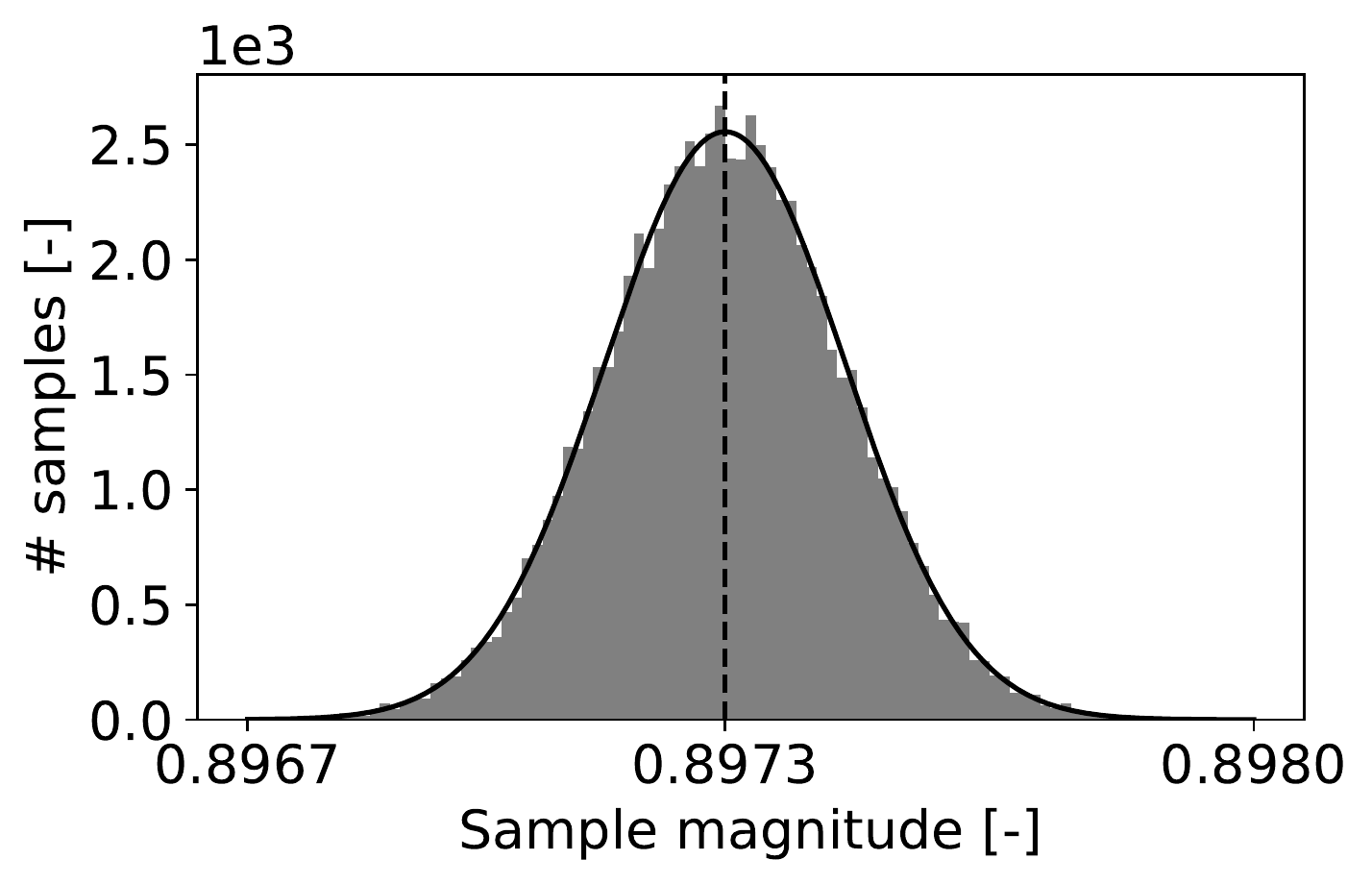}
\subcaption{$n_m=64$}
\label{fig:sampling_mean_posterior_64}
\end{subfigure}
\begin{subfigure}[c]{0.24\textwidth}
\centering
\includegraphics[width=\textwidth]{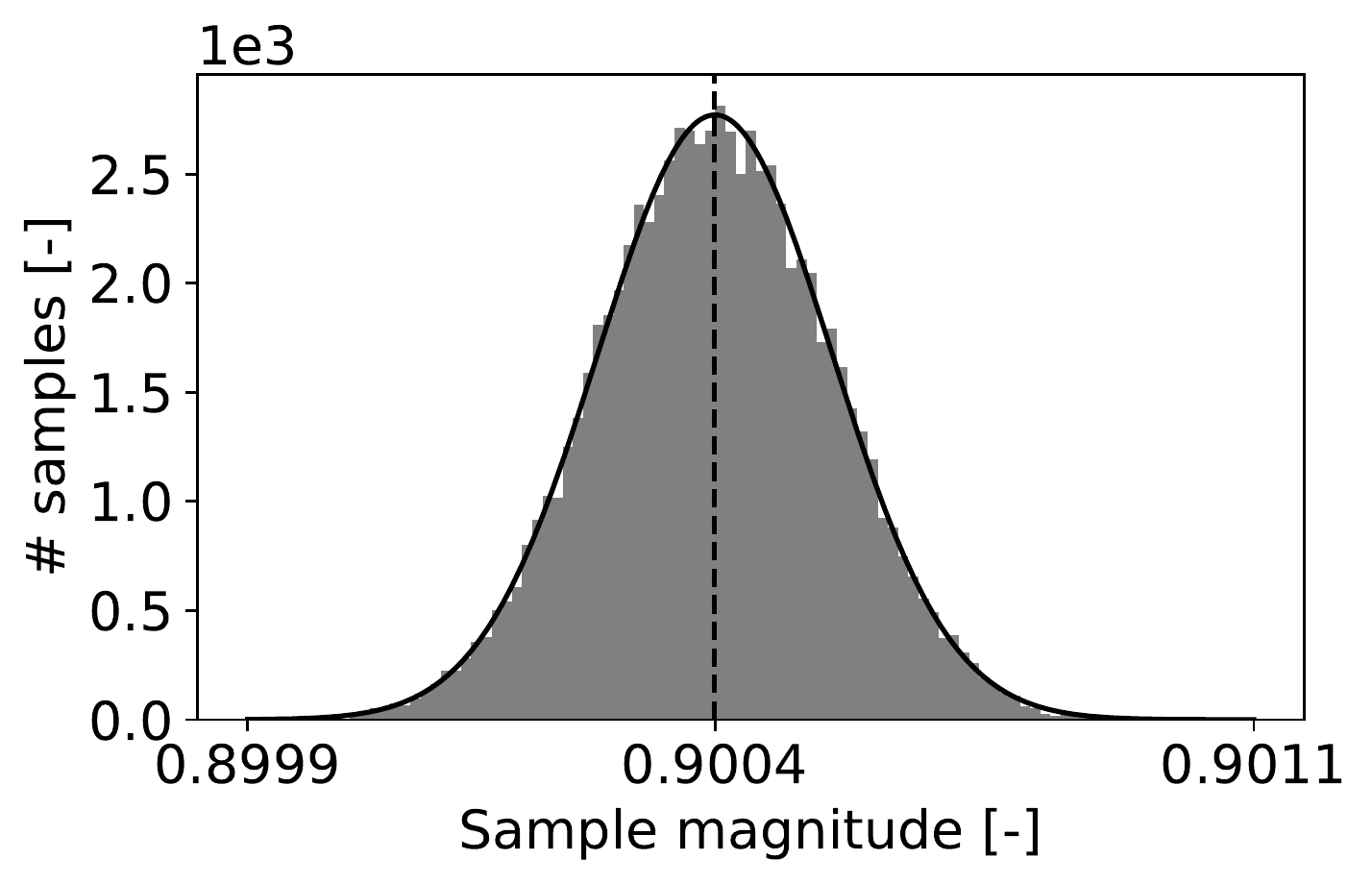}
\subcaption{$n_m=128$}
\label{fig:sampling_mean_posterior_128}
\end{subfigure}
\caption{Sampled state values and a fitted Gaussian probability density function of a single hidden state $\vek{h}_{i,j}$, selected by its largest magnitude for all widths individually. Added is a dashed vertical line representing the propagated mean.}
\label{fig:sampling_mean_posterior}
\end{figure}

Storage expenses may be further reduced by having a non-sampling approach. To avoid sampling, one may focus only on the mean-based estimates. However, as ARD-LSTM network is characterized by nonlinear activation functions, this may cause a problem as $\mathbb{E}\left(f(x)\right) \neq f(\mathbb{E}(x))$. Therefore, sampling is required unless the variance of $f(x)$ is very small and can be neglected. In such a case one may claim $\mathbb{E}(f(x)) \approx f(\mathbb{E}(x))$. As a direct result, sampling can be avoided and the computational efficiency increased while maintaining sufficient accuracy. This is supported by \reffig{fig:sampling_mean_posterior} in which the sampled posterior of the hidden state with the largest magnitude at $t=t_{end}$ and $\epsilon=\SI{0}{\milli\meter}$ is shown. The mean-based (i.e.~no sampling) propagation is depicted as a dashed vertical line.
According to these results one may conclude that the sampling based posterior is a symmetric distribution quite similar to a Gaussian. Its mean is matching the mean obtained by a purely mean-based approach.  

The comparison of the computational efficiency of both the point estimate LSTM and ARD-LSTM is given in \reftab{tab:opti_metrics}. The times per epoch for the propagated mean-based are respectively $\left\lbrace \SI{0.185}{\second}, \SI{0.253}{\second}, \SI{0.562}{\second}, \SI{2.63}{\second}\right\rbrace $ for $n_m \in \left\lbrace 16, 32, 64, 128\right\rbrace $. Given these results one may conclude that the mean-based approach is reducing the simulation time per epoch from \SI{5.6}{\percent} to \SI{18.3}{\percent} without reducing the predicitability significantly. 

\section{Conclusion and discussion}
The objective of this paper was to implement an efficient sparse Bayesian learning framework in the popular recurrent LSTM neural network, applicable in regression driven applications within a pre-defined parameter space. By applying an efficient relevance determination scheme one may easily obtain the proper LSTM architecture given the training data set. 

On a structural application example it has been shown that the ARD-LSTM framework is very well capable to describe the given data set by adapting its complexity accordingly. The computational expenses may have increased compared to the point estimate LSTM, but this is largely compensated by the faster convergence of ARD-LSTM in optimization. To increase the computational efficiency several assumptions have been made that may affect its general applicability. Forward propagation of the predictive mean only should be handled with care and is only valid for relatively small predictive variances. Additionally, the ARD framework does not solve any under-fitting and may in this case cause a noisy optimization. Therefore, ensuring that the network is able to describe the problem complexity remains of great importance.

With the prescribed boundary conditions a promising and computationally efficient \\ self-regulating LSTM network has been created, which does not require prior knowledge of a suitable network architecture and size, while ensuring satisfying data fitment at reasonable computational cost.

\bibliography{bibliography}

\end{document}